\documentclass[11pt]{article}

\usepackage[abbrvbib, preprint]{jmlr2e}
\usepackage[preprint]{jmlr2e}

\usepackage{amsmath,amssymb,amsfonts}
\usepackage{algorithm}
\usepackage{algpseudocode}
\usepackage{graphicx}
\usepackage{textcomp}
\usepackage{xcolor}
\usepackage{blkarray}

\graphicspath{ {./python/} }

\usepackage{hyperref}
\hypersetup{
    colorlinks=true,
    linkcolor=blue,
    filecolor=magenta,      
    urlcolor=cyan,
    pdftitle={Overleaf Example},
    pdfpagemode=FullScreen,
    }

\usepackage{timestamp, todonotes}

\usepackage{lastpage}
\jmlrheading{23}{2023}{1-\pageref{LastPage}}{1/21; Revised 5/22}{9/22}{21-0000}{Hershey, Paffenroth, Pathak, and Tavener}

\ShortHeadings{Recurrent and Non-Recurrent Neural Networks: A discrete dynamical systems approach}{Hershey, Paffenroth, Pathak, and Tavener}
\firstpageno{1}

\newcommand\Wr[1]{\colorbox{magenta}{$#1$}}
\newcommand\Wg[1]{\colorbox{green}{$#1$}}
\newcommand\Wb[1]{\colorbox{cyan}{$#1$}}
\newcommand\Wy[1]{\colorbox{yellow}{$#1$}}

\newcommand{\mv}[1]{\mathbf{#1}}

\begin{document}

\title{Rethinking the Relationship between Recurrent and Non-Recurrent Neural Networks:  A Study in Sparsity}


\author{\name Quincy Hershey \email qbhershey@wpi.edu \\
       \addr Data Science\\
       Worcester Polytechnic Institute\\
       Worcester, MA 01609, USA
       \AND
       \name Randy Paffenroth \email rcpaffenroth@wpi.edu \\
       \addr Mathematical Sciences, Computer Science, and Data Science\\
       Worcester Polytechnic Institute\\
       Worcester, MA 01609, USA
       \AND
       \name Harsh Pathak \email hnpathak@wpi.edu \\
       \addr Data Science\\
       Worcester Polytechnic Institute\\
       Worcester, MA 01609, USA
       \AND
       \name Simon Tavener \email tavener@math.colostate.edu  \\
       \addr Mathematics\\
       Colorado State University\\
       Fort Collins, CO 80523}

\editor{My editor}

\maketitle
\tableofcontents

\begin{abstract} \label{sec:abstract}
    Neural networks (NN) can be divided into two broad categories, recurrent and non-recurrent.  Both types of neural networks are popular and extensively studied, but they are often treated as distinct families of machine learning algorithms.  In this position paper, we argue that there is a closer relationship between these two types of neural networks than is normally appreciated.  We show that many common neural network models, such as Recurrent Neural Networks (RNN), Multi-Layer Perceptrons (MLP), and even deep multi-layer transformers, can all be represented as iterative maps.  

    The close relationship between RNNs and other types of NNs should not be surprising.  In particular, RNNs are known to be Turing complete, and therefore capable of representing any computable function (such as any other types of NNs), but herein we argue that the relationship runs deeper and is more practical than this.  For example, RNNs are often thought to be more difficult to train than other types of NNs, with RNNs being plagued by issues such as vanishing or exploding gradients.  However, as we demonstrate in this paper, MLPs, RNNs, and many other NNs lie on a continuum, and this perspective leads to several insights that illuminate both theoretical and practical aspects of NNs.
    
\end{abstract}


\section{Introduction} \label{sec:introduction}

This paper recasts the concept of Neural Networks (NN) as iterative maps in order to study the underlying principles and assumptions embodied in traditional NN architectures.  Traditional feed-forward multi-layer perceptron (MLP) \cite{lecun2015deep} architectures are typically presented as distinct from recurrent neural network (RNN) \cite{rnn} architectures which in-turn have a reputation of being difficult to train \cite{pmlr-v28-pascanu13, Sherstinsky_2020_rnn, schafer2006recurrent}.  Contrary to this perspective, our research shows that every MLP can be recast as an RNN and much can be learned about RNN architectures by reference to MLP models.  This result is consequential since the universality of RNNs has been previously demonstrated \cite{schafer2006recurrent} and the Turing Completeness of RNN architectures has been understood since the 1990s \cite{siegelmann1991turing}.  Beyond such theoretical considerations, we show how many common NNs can be represented as iterative maps which we define in Definition \ref{defn:block_iterative_map}.  Taking an iterative map perspective for MLPs and many other NN architectures leads to several insights that illuminate both theoretical and practical aspects of NNs. 

The iterative map perspective proposed here both allows us to prove the equivalence described above and inspires us to study NNs from a \emph{discrete dynamical systems} perspective.  As we will demonstrate in what follows, such a perspective can elucidate many aspects of NN architectures.  In particular, such ideas are not only of theoretical interest, but they also lead to improvements in NN implemention and training.  

In this paper we develop notation that provides a context for describing many NN architectures as iterative maps. 
This notation is introduced in \S \ref{sec:notation} and iterated maps, in our context, are defined in Algorithm \ref{alg:INN}.  Of course, many RNNs are inherently iterative maps, and we remind the reader of the iterative structure of RNNs in \S \ref{sec:RNN}.  It is less obvious that MLPs can also be described as iterative maps, and we derive such equations in \S \ref{sec:MLP}.  The results in \S \ref{sec:RNN} and \S \ref{sec:MLP} allow both RNNs and MLPs to be concisely formulated as dynamical systems in \S \ref{sec:dynamical-systems}. We explore extensions of these simple dynamical systems in \S \ref{sec:nearby-infinite-impulse}. Beyond a theoretical nicety, we observe that MLPs and many other NNs can \emph{efficiently} be implemented as iterative maps, and we provide a description of an implementation we call Sequential2D in \S \ref{sec:implementation}. 
In \S \ref{sec:numerical-experiments}, we demonstrate that the iterative map perspective for MLPs can lead to surprising numerical results.  In particular, we show that random NN architectures are equivalent, in the appropriate sense, to more standard layer-wise approaches.
Further exploration of new classes of methods that are natural extensions to existing methods will be the subject of future publications.


Deep networks have also been considered as discretizations of continuous dynamical systems and training viewed as an optimal control problem.  This perspective has created an enormous literature and an exhaustive description is challenging. As an introduction to this area, we point to the early work of 
\cite{chen2018neural, weinan2017proposal, chang2017multi, han2019mean, gunther2020layer}.  An early review appears in \cite{liu2019deep}.
Also, dynamical systems already play several important roles in state-of-the-art NNs as can be seen in reviews such as \cite{liu2019deep}.  However, several specific parts of the current literature bear special attention.  For example, Koopman operators have an important role to play in analyzing NNs \cite{yeung2019learning}.  Also, several recent state-of-the-art models are seemingly inspired by ideas similar to those that inspire our work, such as diffusion models \cite{croitoru2023diffusion,yang2023diffusion} and structured state space models \cite{gu2021efficiently} such as MAMBA \cite{gu2023mamba}.  Finally, iterated auto-encoders and their interpretation in terms of associative memory \cite{radhakrishnan2020overparameterized} is quite close in spirit to the ideas that we present here.  It it our hope that our contributions in this paper serve to further underscore the important connections between NNs and dynamical systems.

\section{Notation} \label{sec:notation}
As function composition is a fundamental building block of NNs and iterative maps in general, it is important to establish a notation for composing functions.  
We begin with the following notational conventions for scalars, vectors and matrices as shown in Table \ref{tab:notation}.

\begin{table}[h]
    \centering
    \begin{tabular}{|c|c|c|}
        \hline
        \textbf{Notation} & \textbf{Description} & \textbf{Example} \\
        \hline
        $x$ & Scalar & $3.14$ \\
        \hline
        $\mv{x}$ & Column vector & $\begin{bmatrix} 1 \\ 2 \\ 3 \end{bmatrix}$ \\
        \hline
        $|\mv{x}|$ & The number of entries in $\mv{x}$ & $\left| \begin{bmatrix} 1 \\ 2 \\ 3 \end{bmatrix} \right| = 3$\\
        \hline
        $X$ & Matrix & $\begin{bmatrix} 1 & 2 \\ 3 & 4 \end{bmatrix}$ \\
        \hline
    \end{tabular}
    \caption{Notation for scalars, vectors, and matrices.}
    \label{tab:notation}
\end{table}

We will use the function composition operator $\circ$ to write

\begin{equation} \label{eq:composition_operator}
    f(g(\mv{x})) = f \circ g \circ \mv{x}
\end{equation}
for \emph{vector valued functions} $f$ and $g$.  To be clear, when we write $f \circ g \circ \mv{x}$ we intend that the argument $\mv{x}$ will only appear as the right-most entry and $f$ and $g$ are functions.  We then slightly abuse notation by defining a $2 \times 2$ \emph{block non-linear function} as
\begin{equation}
    \begin{bmatrix}
        f_{0,0} & f_{0,1} \\
        f_{1,0} & f_{1,1} \\
    \end{bmatrix}
    \circ
    \begin{bmatrix}
        \mv{x} \\
        \mv{h} \\
    \end{bmatrix}
    =
    \begin{bmatrix}
        f_{0,0}(\mv{x}) + f_{0,1}(\mv{h}) \\
        f_{1,0}(\mv{x}) + f_{1,1}(\mv{h}) \\
    \end{bmatrix}.
    \label{eq:composition_operator_blockwise}
\end{equation}

It's important to emphasize that the block non-linear function defined in equation \eqref{eq:composition_operator_blockwise} is not a matrix; instead, it's a separable function. This function operates on a vector comprising two input vectors and produces a vector consisting of two output vectors. Also, it's worth noting that the choice of combining the outputs of $f_{i,0}$ and $f_{i,1}$ isn't restricted to addition alone. However, for the sake of simplicity and to maintain resemblance to standard block matrix equations, addition is employed in this paper.

Moreover, if the functions $f_{i,j}$ are linear, the block non-linear function in equation \eqref{eq:composition_operator_blockwise} simplifies to a block matrix equation. In essence, for linear maps, the function composition operator and the dot product are equivalent.

\begin{definition} \label{defn:block-nonlinear-function}
Given functions $f_{i,j}$, where each $f_{i,j}$ may be linear or non-linear, we define a \emph{block non-linear function} as 
\begin{equation} \label{eq:block-nonlinear-function}
    \begin{aligned}
        &F = \\
        &\begin{bmatrix}
            f_{0,0}: \mathbb{R}^{|\mv{x}|}\xrightarrow[]{}\mathbb{R}^{|\mv{x}|} &
            \hdots &
            f_{0,n-1}: \mathbb{R}^{|\mv{h}_k|}\xrightarrow[]{}\mathbb{R}^{|\mv{x}|} &
            f_{0,n}: \mathbb{R}^{|\mv{y}|}\xrightarrow[]{}\mathbb{R}^{|\mv{x}|} \\
            f_{1,0}: \mathbb{R}^{|\mv{x}|}\xrightarrow[]{}\mathbb{R}^{|\mv{h}_1|} &
            \hdots &
            f_{1,n-1}: \mathbb{R}^{|\mv{h}_k|}\xrightarrow[]{}\mathbb{R}^{|\mv{h}_1|} &
            f_{1,n}: \mathbb{R}^{|\mv{y}|}\xrightarrow[]{}\mathbb{R}^{|\mv{h}_1|} \\
            \vdots & \ddots & \vdots & \vdots \\
            f_{m-1,0}: \mathbb{R}^{|\mv{x}|}\xrightarrow[]{}\mathbb{R}^{|\mv{h}_k|} &
            \hdots &
            f_{m-1,n-1}: \mathbb{R}^{|\mv{h}_k|}\xrightarrow[]{}\mathbb{R}^{|\mv{h}_k|} &
            f_{m-1,n}: \mathbb{R}^{|\mv{y}|}\xrightarrow[]{}\mathbb{R}^{|\mv{h}_k|} \\
            f_{m,0}: \mathbb{R}^{|\mv{x}|}\xrightarrow[]{}\mathbb{R}^{|\mv{y}|} &
            \hdots &
            f_{m,n-1}: \mathbb{R}^{|\mv{h}_k|}\xrightarrow[]{}\mathbb{R}^{|\mv{y}|} &
            f_{m,n}: \mathbb{R}^{|\mv{y}|}\xrightarrow[]{}\mathbb{R}^{|\mv{y}|} \\
        \end{bmatrix}
    \end{aligned} \, .
\end{equation} 
The action of $F$ on a vector is then defined as 
\begin{equation} \label{eq:block-nonlinear-function-action}
    \begin{aligned}
        F
        \circ
        \begin{bmatrix}
            x \in \mathbb{R}^{|\mv{x}|} \\
            h_1 \in \mathbb{R}^{|\mv{h}_1|} \\
            \vdots \\
            h_k \in \mathbb{R}^{|\mv{h}_k|} \\
            y \in \mathbb{R}^{|\mv{y}|}
        \end{bmatrix}&=
        \begin{bmatrix}
        f_{0,0}(\mv{x}) + f_{0,1}(\mv{h}_1) + \hdots + f_{0,n-1}(\mv{h}_k) + f_{0,n}(\mv{y}) \\ 
        f_{1,0}(\mv{x}) + f_{1,1}(\mv{h}_1) + \hdots + f_{1,n-1}(\mv{h}_k) + f_{1,n}(\mv{y}) \\
        \vdots \\
        f_{m-1,0}(\mv{x}) + f_{m-1,1}(\mv{h}_1) + \hdots + f_{m-1,n-1}(\mv{h}_k) + f_{m-1,n}(\mv{y}) \\ 
        f_{m,0}(\mv{x}) + f_{m,1}(\mv{h}_1) + \hdots + f_{m,n-1}(\mv{h}_k) + f_{m,n}(\mv{y}) \\ 
        \end{bmatrix} \\
        &\in \mathbb{R}^{|\mv{x}| + |\mv{h}_1| + \hdots + |\mv{h}_k| + |\mv{y}|} \, ,
    \end{aligned}
\end{equation} 
\end{definition}        
thus in many ways, \eqref{eq:block-nonlinear-function} is a straight forward generalization of a block matrix.  

When working with NNs, training algorithms are often considered in terms of \emph{mini-batches}, where a set of $n$ training examples are processed together.  
\footnote{In NN literature, parameter matrices are often applied to a mini-batch using a left dot-product, such as $X \cdot W$, where $X$ is a mini-batch of training examples and $W$ is a parameter matrix.  In this paper, we use the right dot-product, such as $W \cdot X$, which is more common in the linear algebra literature and matches the function composition notation $f \circ g(x)$.}

\begin{definition} \label{defn:block_linear_function_minibatch}
Given matrices $X \in \mathbb{R}^{|\mv{x}| \times n}, H_1 \in \mathbb{R}^{|\mv{h}_1| \times n}, \cdots, H_k \in \mathbb{R}^{|\mv{h}_k| \times n}, Y \in \mathbb{R}^{|\mv{y}| \times n}$ then the block non-linear function $F$ acts on the mini-batch as
\begin{equation} \label{eq:block_linear_function_minibatch}
    F
    \circ
    \begin{bmatrix}
         X \in \mathbb{R}^{|\mv{x}| \times n} \\
        H_1 \in \mathbb{R}^{|\mv{h}_1| \times n} \\
        \vdots \\
         H_k \in \mathbb{R}^{|\mv{h}_k| \times n} \\
         Y \in \mathbb{R}^{|\mv{y}| \times n}
    \end{bmatrix} =
     \begin{bmatrix}
        f_{0,0}(X) + f_{0,1}(H_1) + \hdots + f_{0,n-1}(H_k) + f_{0,n}(Y) \\ 
        f_{1,0}(X) + f_{1,1}(H_1) + \hdots + f_{1,n-1}(H_k) + f_{1,n}(Y) \\
        \vdots \\
        f_{m-1,0}(X) + f_{m-1,1}(H_1) + \hdots + f_{m-1,n-1}(H_k) + f_{m-1,n}(Y) \\ 
        f_{m,0}(X) + f_{m,1}(H_1) + \hdots + f_{m,n-1}(H_k) + f_{m,n}(Y) \\ 
    \end{bmatrix}
\end{equation}
where each function $f_{i,j}$ acts \emph{columnwise} on the given input matrix. This gives rise to
\begin{equation} \label{block_linear_function_minibatch_action}
    F
    \circ
    \begin{bmatrix}
        X \in \mathbb{R}^{|\mv{x}| \times n} \\
        H_1 \in \mathbb{R}^{|\mv{h}_1| \times n} \\
        \vdots \\
        H_k \in \mathbb{R}^{|\mv{h}_k| \times n} \\
        Y \in \mathbb{R}^{|\mv{y}| \times n}
    \end{bmatrix} \in \mathbb{R}^{(|\mv{x}| + |\mv{h}_1| + \hdots + |\mv{h}_k| + |\mv{y}|) \times n} \, .
    \end{equation}
\end{definition}

In what follows we will often wish to iterate a \emph{fixed} block non-linear function $F$.  In particular, a defining characteristic of a discrete iterated map is that the map of interest is fixed and applied a number of times to a given input $x$ as in
$f(f(\cdots(f(x))\cdots))$.  Accordingly, we define an iterated block nonlinear function as below.

\begin{definition}\label{defn:block_iterative_map}
An iterative map is defined below. 
\begin{algorithm}
\caption{INN} \label{alg:INN}
\begin{algorithmic}
\State $t \gets 0$
\State $\mv{x_0} \gets \mv{x}$
  \While {$t<T$}
    \State $\mv{x}_{t+1} \gets F \; \circ \; \mv{x}_{t}$
    \State $t \gets t+1$
  \EndWhile
\end{algorithmic}
\end{algorithm}
\end{definition}


\section{Recurrent Neural Networks (RNNs)} \label{sec:RNN}

\subsection{Traditional development of RNNs}

A recurrent neural network (RNN) is a type of neural network that is designed to process sequential data, e.g., data that is indexed by time. Let $\mv{h}_k \in \mathbb{R}^n$ be the \emph{state} of the system, and $\mv{x}_k \in \mathbb{R}^m$ be the input data or forcing function.  As in \cite{Goodfellow-et-al-2016, pmlr-v28-pascanu13}, an RNN can be written as 
\begin{equation} \label{eq:RNN-defn}
\begin{aligned}
\hbox{Given}\;  &\mv{h}_0 = \mv{x}_0 \\
&\mv{h}_{t+1} = f_{\theta} (\mv{x}_{t+1}, \mv{h}_t), \quad t=0,1,\dots,T-1 \\
\end{aligned}
\end{equation}
where $f_\theta $ is a family of functions parameterized by $\theta$. In other words the $\mv{x}_0$ is the initialization of the hidden state $\mv{h}_0$.  Equivalently, equation \eqref{eq:RNN-defn} defines a discrete time dynamical system for $\mv{h}_k$ where the function $f$ is the \emph{dynamics} of the system and $\mv{x}_k$ is a forcing term.  The study of RNNs revolves around the choice of the function family $f_\theta(\cdot, \cdot)$, with many interesting sub-classes of RNNs having different properties. 


A specific family of functions that is widely used in the RNN literature \cite{Goodfellow-et-al-2016, pmlr-v28-pascanu13, salehinejad2017recent, lipton2015critical} is a family of affine functions with non-linear activations, e.g.,

\begin{equation} \label{eq:RNN-activation-function}
\mv{h}_{t+1} = \sigma(W_x \cdot \mv{x}_{t+1} + W_h \cdot \mv{h}_t + \mv{b})
\end{equation}
where we have the weight matrices $W_x \in \mathbb{R}^{n \times m}$ and $W_h \in \mathbb{R}^{n \times n}$, the bias vector $\mv{b} \in \mathbb{R}^n$, and $\sigma$ is a non-linear activation function, most commonly a {\it tanh}.   

\subsection{RNNs as iterated maps} \label{sec:RNN-iterative}

The RNN in equation \eqref{eq:RNN-activation-function} can be written as a block non-linear function 
\begin{equation} \label{eq:RNN-activation-function-block}
\begin{aligned}
    \begin{bmatrix}
        0 & 0 \\
        0 & \sigma\\
    \end{bmatrix}
    \circ
    \begin{bmatrix}
        0 & 0 \\
        F_x & F_h\\
    \end{bmatrix}
    \circ
    \begin{bmatrix}
        \mv{x}_{t+1}     \\
        \mv{h}_{t} \\
    \end{bmatrix}
    & =
    \begin{bmatrix}
        0 \\
        \sigma( F_x(\mv{x}_{t+1}) + F_h(\mv{h}_t) )\\
    \end{bmatrix} 
\end{aligned}
\end{equation}
where $F_x(\mv{x}_{t+1}) = W_x \cdot \mv{x}_{t+1}$ is a linear function, $F_h(\mv{h}_{t}) = W_h \cdot \mv{h}_t + \mv{b}$ is an affine function and where $\sigma:\mathbb{R}^{n} \mapsto \mathbb{R}^{n}$.  In general however, $F_x$ and $F_h$ may be nonlinear functions such that $F_x: \mathbb{R}^{m} \mapsto \mathbb{R}^{n}$ and $F_h: \mathbb{R}^{n} \mapsto \mathbb{R}^{n}$, but the choice in  \eqref{eq:RNN-activation-function} is common in the RNN literature \cite{Goodfellow-et-al-2016, pmlr-v28-pascanu13, salehinejad2017recent, lipton2015critical}. 

The trainable parameter set is
\begin{equation} \label{eq:RNN-parameter-set}
\theta_{RNN} = \{W_x, W_h, \mv{b} \}.    
\end{equation}
RNNs can therefore be viewed as a forced discrete dynamical system with state space $h_k$
 and the function
 \begin{equation} \label{eq:RNN-dyamics}
M_{RNN} = 
 \begin{aligned}
    \begin{bmatrix}
        0 & 0 \\
        0 & \sigma \\
    \end{bmatrix}
    \circ
    \begin{bmatrix}
        0 & 0 \\
        F_x & F_h\\
    \end{bmatrix}
\end{aligned}    
\end{equation}
is the \emph{dynamics} of the system.  
Given $\mv{h}_0$, the sequence provided by \eqref{eq:RNN-activation-function}, or equivalently \eqref{eq:RNN-dyamics}, is
\begin{equation} \label{eq:RNN-sequence-3}
\begin{aligned}
        \mv{h}_1 &= \sigma ( F_x(\mv{x}_1) + F_h(\mv{h}_0) ) \\
        \mv{h}_2 &= \sigma \Big( F_x(\mv{x}_2) + F_h(\mv{h}_1) \Big) \\
                 &= \sigma \Big( F_x(\mv{x}_2) 
                               + F_h \big( \; \sigma( F_x(\mv{x}_1) + F_h(\mv{h}_0) ) \; \big) \Big) \\
        \mv{h}_3 &= \sigma \Bigg( F_x(\mv{x}_3) + F_h(\mv{h}_2) \Bigg) \\     
                 &= \sigma \Bigg( F_x(\mv{x}_3) 
                   + F_h \bigg( \;\; \sigma \Big( F_x(\mv{x}_2) 
                     + F_h \big( \; \sigma( F_x(\mv{x}_1) + F_h(\mv{h}_0)) \; \big) \Big) \;\; \bigg) \Bigg) \, .
\end{aligned} 
\end{equation}

To emphasize the iterative nature of the RNN, we extend the definition of the RNN in \eqref{eq:RNN-activation-function} by defining
\begin{equation} \label{eq:RNN-ftheta}
    f_{\theta}(\mv{x},\mv{h})= \sigma(F_x(\mv{x}) + F_h(\mv{h}))
\end{equation}
and using this notation and appealing to the standard $\circ$ notation for function composition we see that 
\begin{equation} \label{eq:RNN-h3}
\begin{aligned}
\mv{h}_3 &=  f_\theta(\mv{x}_3,\cdot) \circ f_\theta(\mv{x}_2,\cdot) \circ f_\theta(\mv{x}_1,\cdot) \circ \mv{h}_0\\
     &=  f_\theta \Big(\mv{x}_3, f_\theta \big(\mv{x}_2, f_\theta(\mv{x}_1,\mv{h}_0)  \big)  \Big) \\
    &=  f_\theta \Big(\mv{x}_3, f_\theta \big(\mv{x}_2, f_\theta(\mv{x}_1,\mv{x}_0)  \big)  \Big)  \, .
\end{aligned} 
\end{equation}

Equivalently, we can iterate the following matrix three times and recover the output of \eqref{eq:RNN-h3} in the final component of the vector.  While this notation appears superfluous, it demonstrates that we create a \emph{fixed} map composed of a \emph{fixed} function $f_\theta$.  This approach will prove to be particularly useful looking ahead to \S \ref{sec:MLP} when we consider MLPs. 
We define 
\begin{equation} \label{eq:RNN-iterative-matrix-3}
M_{RNN3}=
\begin{bmatrix}
    0                         & 0                        & 0                        & 0\\
    f_\theta(\mv{x}_1,\cdot)  & 0                        & 0                        & 0\\
    0                         & f_\theta(\mv{x}_2,\cdot) & 0                        & 0\\ 
    0                         & 0                        & f_\theta(\mv{x}_3,\cdot) & 0\\
\end{bmatrix}.
\end{equation}
We show below that we reproduce the action of \eqref{eq:RNN-dyamics} by expanding the dimension of the space such all copies of \eqref{eq:RNN-dyamics} lie ``below the diagonal''.
This is an example of an unrolling operation utilizing an \emph{increase in dimension}.  Why do we combine these two operations?  As we will see later by combining these two operations we will uncover a unifying representation of RNNs and MLPs.  In the general RNN case, the particular dimension we chose for the lifting is $\big( |\mv{h}| + T(|\mv{x}| + |\mv{h}|) \big)$ where $T$ is the number of times the map is applied to produce $\mv{h}_T$, and $|\mv{x}| = |\mv{x}_i| \; \forall \; i=1,\dots, T-1$ and $|\mv{h}| = |\mv{h}_i| \; \forall \; i=0,\dots, T$. 
\footnote{In the context of MLPs we will use a similar lifting approach to create a \emph{fixed} map, but in this case the fixed map will be composed of \emph{different} functions,  $f_{\theta_1},f_{\theta_2}, \dots, f_{\theta_T}$.  In general, these functions map from and to vector spaces having different dimensions as indicated in \eqref{eq:block-nonlinear-function}.  The dimension of the fixed map in the MLP case is therefore $|\mv{h}_0| + |\mv{h}_1| + |\mv{h}_2| + \dots + |\mv{h}_T|$.}

Given the initial state 
\begin{equation} \label{eq:RNN-iterative-initial-condition-3}
    \begin{bmatrix}
        \mv{h}_0 \\
        0 \\
        0 \\
        0 \\
    \end{bmatrix} 
\end{equation}
the first iteration of the map $M_{RNN3}$ (abbreviated to $M$ for convenience) gives
\begin{equation} \label{eq:RNN-iteration-1}
    M \circ 
    \begin{bmatrix}
        \mv{h}_0 \\
        0 \\
        0 \\
        0 \\
    \end{bmatrix} 
     =  
    \begin{bmatrix}
        0                             & 0                             & 0                             & 0 \\
        \Wg{f_\theta(\mv{x}_1,\cdot)} & 0                             & 0                             & 0 \\
        0                             & \Wg{f_\theta(\mv{x}_2,\cdot)} & 0                             & 0 \\
        0                             & 0                             & \Wg{f_\theta(\mv{x}_3,\cdot)} & 0 \\
    \end{bmatrix}
        \circ
    \begin{bmatrix}
        \mv{h}_0 \\
        0 \\
        0 \\
        0 \\
    \end{bmatrix} 
    = 
    \begin{bmatrix}
        0 \\
        \Wg{f_\theta(\mv{x}_1,\mv{h}_0)} \\
        \Wg{f_\theta(\mv{x}_2,0)} \\
        \Wg{f_\theta(\mv{x}_3,0)} \\
    \end{bmatrix}.
\end{equation}
The second iteration of the map $M$ gives
\begin{equation} \label{eq:RNN-iteration-2}
\begin{aligned}
    M \circ 
    M \circ 
    \begin{bmatrix}
        \mv{h}_0 \\
        0 \\
        0 \\
        0 \\
    \end{bmatrix} 
    &= 
    \begin{bmatrix}
        0                             & 0                             & 0                             & 0 \\
        \Wg{f_\theta(\mv{x}_1,\cdot)} & 0                             & 0                             & 0 \\
        0                             & \Wg{f_\theta(\mv{x}_2,\cdot)} & 0                             & 0 \\
        0                             & 0                             & \Wg{f_\theta(\mv{x}_3,\cdot)} & 0 \\
    \end{bmatrix}
    \circ 
    \begin{bmatrix}
        0 \\
        \Wg{f_\theta(\mv{x}_1,\mv{h}_0)} \\
        \Wg{f_\theta(\mv{x}_2,0)} \\
        \Wg{f_\theta(\mv{x}_3,0)} \\
    \end{bmatrix} \\
    &=
    \begin{bmatrix}
        0 \\
        \Wg{f_\theta(\mv{x}_1,0)} \\
        \Wg{f_\theta\big( \mv{x}_2, ( f_\theta(\mv{x}_1,\mv{h}_0) \big)} \\
        \Wg{f_\theta\big( \mv{x}_3, ( f_\theta(\mv{x}_2, 0) \big)} \\
    \end{bmatrix}   
\end{aligned}
\end{equation}
and the third iteration of the map $M$ results in
\begin{equation} \label{eq:RNN-iteration-3}
\begin{aligned}
    M \circ 
    M \circ 
    &M \circ 
    \begin{bmatrix}
        \mv{h}_0 \\
        0 \\
        0 \\
        0 \\
    \end{bmatrix} \\
    &= 
    \begin{bmatrix}
        0                             & 0                             & 0                             & 0 \\
        \Wg{f_\theta(\mv{x}_1,\cdot)} & 0                             & 0                             & 0 \\
        0                             & \Wg{f_\theta(\mv{x}_2,\cdot)} & 0                             & 0 \\
        0                             & 0                             & \Wg{f_\theta(\mv{x}_3,\cdot)} & 0 \\
    \end{bmatrix}
    \circ 
    \begin{bmatrix}
        0 \\
        \Wg{f_\theta(\mv{x}_1,0)} \\
        \Wg{f_\theta\big( \mv{x}_2, ( f_\theta(\mv{x}_1,\mv{h}_0) \big)} \\
        \Wg{f_\theta\big( \mv{x}_3, ( f_\theta(\mv{x}_2, 0) \big)} \\
    \end{bmatrix}      \\
    &=
    \begin{bmatrix}
        0 \\
        \Wg{f_\theta(\mv{x}_1,0)} \\
        \Wg{f_\theta \big( \mv{x}_2, f_\theta(\mv{x}_1,0) \big)} \\
        \Wg{f_\theta \Big(\mv{x}_3,  f_\theta \big(\mv{x}_2, f_\theta(\mv{x}_1,\mv{h}_0) \big) \Big)}  \\
    \end{bmatrix} \\
    &=
    \begin{bmatrix}
        0 \\
        \Wg{f_\theta(\mv{x}_1,0)} \\
        \Wg{f_\theta \big( \mv{x}_2, f_\theta(\mv{x}_1,0) \big)} \\
        \Wg{f_\theta \Big(\mv{x}_3,  f_\theta \big(\mv{x}_2, f_\theta(\mv{x}_1,\mv{x}_0) \big) \Big)}  \\
    \end{bmatrix}   \, .
\end{aligned} 
\end{equation}
If we now interpret the last entry in the vector as the output of the map, we observe that it is \emph{identical} to the output of a three layer RNN with function/dynamics
$\Wg{f_\theta(\cdot,\cdot)}$, forcing terms $\mv{x}_1,\mv{x}_2,$ and $\mv{x}_3$,
and input $\mv{x}_0$ as shown in \eqref{eq:RNN-h3}. 

\subsection{Choice of the initial vector} \label{sec:RNN3-q}

Given the initial state 
\begin{equation} \label{eq:RNN-iterative-initial-condition-q}
    \begin{bmatrix}
        \mv{h}_0 \\
        q_1 \\
        q_2 \\
        q_3 \\
    \end{bmatrix}
\end{equation}
the first iteration of the map $M_{RNN3}$ (abbreviated to $M$ for convenience) gives
\begin{equation} \label{eq:RNN-iteration-1q}
    M \circ 
    \begin{bmatrix}
        \mv{h}_0 \\
        q_1 \\
        q_2 \\
        q_3 \\
    \end{bmatrix} 
     =  
    \begin{bmatrix}
        0                             & 0                             & 0                             & 0 \\
        \Wg{f_\theta(\mv{x}_1,\cdot)} & 0                             & 0                             & 0 \\
        0                             & \Wg{f_\theta(\mv{x}_2,\cdot)} & 0                             & 0 \\
        0                             & 0                             & \Wg{f_\theta(\mv{x}_3,\cdot)} & 0 \\
    \end{bmatrix}
        \circ
    \begin{bmatrix}
        \mv{h}_0 \\
       q_1 \\
        q_2 \\
        q_3 \\
    \end{bmatrix} 
    = 
    \begin{bmatrix}
        0 \\
        \Wg{f_\theta(\mv{x}_1,\mv{h}_0)} \\
        \Wg{f_\theta(\mv{x}_2,q_1)} \\
        \Wg{f_\theta(\mv{x}_3,q_2)} \\
    \end{bmatrix} \, .
\end{equation}
The second iteration of the map $M$ gives
\begin{equation} \label{eq:RNN-iteration-2q}
\begin{aligned}
    M \circ 
    M \circ 
    \begin{bmatrix}
        \mv{h}_0 \\
        q_1 \\
        q_2 \\
        q_3 \\
    \end{bmatrix} 
    &= 
    \begin{bmatrix}
        0                             & 0                             & 0                             & 0 \\
        \Wg{f_\theta(\mv{x}_1,\cdot)} & 0                             & 0                             & 0 \\
        0                             & \Wg{f_\theta(\mv{x}_2,\cdot)} & 0                             & 0 \\
        0                             & 0                             & \Wg{f_\theta(\mv{x}_3,\cdot)} & 0 \\
    \end{bmatrix}
    \circ 
    \begin{bmatrix}
        0 \\
        \Wg{f_\theta(\mv{x}_1,\mv{h}_0)} \\
        \Wg{f_\theta(\mv{x}_2,q_1)} \\
        \Wg{f_\theta(\mv{x}_3,q_2)} \\
    \end{bmatrix} \\
    &=
    \begin{bmatrix}
        0 \\
        \Wg{f_\theta(\mv{x}_1,0)} \\
        \Wg{f_\theta\big( \mv{x}_2, ( f_\theta(\mv{x}_1,\mv{h}_0) \big)} \\
        \Wg{f_\theta\big( \mv{x}_3, ( f_\theta(\mv{x}_2, q_1) \big)} \\
    \end{bmatrix}   
\end{aligned}
\end{equation}
and the third iteration of the map $M$ results in
\begin{equation} \label{eq:RNN-iteration-3q}
\begin{aligned}
    M \circ 
    M \circ 
    &M \circ 
    \begin{bmatrix}
        \mv{h}_0 \\
        q_1 \\
        q_2 \\
        q_3 \\
    \end{bmatrix} \\
    &= 
    \begin{bmatrix}
        0                             & 0                             & 0                             & 0 \\
        \Wg{f_\theta(\mv{x}_1,\cdot)} & 0                             & 0                             & 0 \\
        0                             & \Wg{f_\theta(\mv{x}_2,\cdot)} & 0                             & 0 \\
        0                             & 0                             & \Wg{f_\theta(\mv{x}_3,\cdot)} & 0 \\
    \end{bmatrix}
    \circ 
    \begin{bmatrix}
        0 \\
        \Wg{f_\theta(\mv{x}_1,0)} \\
        \Wg{f_\theta\big( \mv{x}_2, ( f_\theta(\mv{x}_1,\mv{h}_0) \big)} \\
        \Wg{f_\theta\big( \mv{x}_3, ( f_\theta(\mv{x}_2, q_1) \big)} \\
    \end{bmatrix}    \\
    &=
    \begin{bmatrix}
        0 \\
        \Wg{f_\theta(\mv{x}_1,0)} \\
        \Wg{f_\theta \big( \mv{x}_2, f_\theta(\mv{x}_1,0) \big)} \\
        \Wg{f_\theta \Big(\mv{x}_3,  f_\theta \big(\mv{x}_2, f_\theta(\mv{x}_1,\mv{h}_0) \big) \Big)} \\
    \end{bmatrix}  \\ 
    &=
    \begin{bmatrix}
        0 \\
        \Wg{f_\theta(\mv{x}_1,0)} \\
        \Wg{f_\theta \big( \mv{x}_2, f_\theta(\mv{x}_1,0) \big)} \\
        \Wg{f_\theta \Big(\mv{x}_3,  f_\theta \big(\mv{x}_2, f_\theta(\mv{x}_1,\mv{x}_0) \big) \Big)} \\
    \end{bmatrix} \, .
\end{aligned}
\end{equation}

The final component after three iterations is equal to \eqref{eq:RNN-iteration-3} and independent of $q_1, q_2,q_3$.
\section{Multilayer perceptrons (MLPs)} \label{sec:MLP}

\subsection{Traditional development of MLPs} \label{sec:MLP-traditional}
Neural networks such as the traditional feed-forward multilayer perceptron \cite{MURTAGH1991183} architecture can be defined as a set of nested functions.  In this representation, each layer $i$ of the network and accompanying (fixed) nonlinear activation function $\sigma(\cdot)$ \cite{9108717} is represented as a function.  Let $\theta_i$ represent the parameters that define $W_i$ and $\mv{b}_i$. 
To make the notation consistent we make  $\mv{h}_0$ equal to the input data $\mv{x}$.\footnote{We note that for MLP the input data appears only once, yet the forcing function in an RNN changes every iteration and $\mv{h}_0 = \mv{x}_0$.} Let

\begin{equation}\label{eq:MLP-activation}
\mv{h}_{i+1} = f_{\theta_{i+1}}(\mv{h}_i) = \sigma(W_{i+1} \cdot \mv{h}_i + \mv{b}_{i+1}), 
\quad i=0,\dots,T-1 
\end{equation}
with $\mv{h}_0 \in \mathbb{R}^m$, $\mv{h}_i \in \mathbb{R}^{|\mv{h}_i|}$, and weight matrix $W_{i} \in \mathbb{R}^{|\mv{h}_{i}| \times |\mv{h}_{i-1}|}$ and bias vector $\mv{b} \in \mathbb{R}^{|\mv{h}_{i}|}$ forming a parameter set 

\begin{equation}\label{eq:MLP-parameters}
    \theta_{MLP} = \{W_i, \mv{b}_i \}, \;i=1, \dots, T.
\end{equation}

Each $h_i, \; i=1,\dots,T$ is thought of as a \emph{hidden layer} of the neural network.  Such a layer is called a \emph{dense layer} based on the density of the non-zero entries in the matrix $W_i$.  Note that the affine function $W_{i} \cdot \mv{h}_{i-1} + \mv{b}_{i}$ of each layer is fed through the nonlinear activation function $\sigma$.  The final layer provides model output $\tilde{\mv{y}} = \mv{h}_T$.  
As a concrete example, consider the following three-layer network, where we nest three functions of the form \eqref{eq:MLP-activation} to obtain
\begin{equation}\label{eq:MLP-traditional-h3}
    \tilde{\mv{y}} = \mv{h}_3 = \sigma \Big( W_3 \cdot \sigma \big( \; W_2 \cdot \sigma( \; W_1 \cdot \mv{h}_0 + \mv{b}_1 \;) + \mv{b}_2 \; \big) + \mv{b}_3 \Big)
\end{equation} 
or, equivalently and more compactly,
\begin{equation}\label{eq:MLP-traditional-compact-h3}
    \tilde{\mv{y}} = \mv{h}_3 = f_{\theta_3}(f_{\theta_2}(f_{\theta_1}(\mv{h}_0)))
\end{equation}
where
\begin{equation}\label{eq:MLP-traditional-compact-definition}
    f_{\theta_{j}}(\mv{z}) = \sigma(W_{j} \cdot \mv{z} + \mv{b}_{j}).
\end{equation}
In general
\begin{equation}\label{eq:MLP-traditional-hT}
\begin{aligned}
\tilde{\mv{y}} = \mv{h}_T &= f_{\theta_T} \circ f_{\theta_{T-1}} \circ \dots \circ f_{\theta_1} \circ \mv{h}_0 \\
 &= f_{\theta_T} \Big(  \;\; f_{\theta_{T-1}} \big( \dots ( \; f_{\theta_1}(\mv{h}_0) \; \big) \;\; \Big) \, .
\end{aligned}
\end{equation}

\subsection{MLPs as iterated maps} \label{sec:MLP-block-iterative}

Given $\mv{h}_0 = \mv{x}$ the iteration \eqref{eq:MLP-traditional-hT} gives
\begin{equation} \label{eq:MLP-iterative-h3}
\begin{aligned}
        \mv{h}_1 &= \sigma( W_1\cdot \mv{h}_0 + \mv{b}_1 ) \\
        \mv{h}_2 &= \sigma \big( W_2 \cdot \mv{h}_1 + \mv{b}_2 \big) \\
            &= \sigma \big( W_2 \cdot \sigma (W_1\cdot \mv{h}_0 + \mv{b}_1) + \mv{b}_2 \big) \\
        \mv{h}_3 &= \sigma \Big( W_3 \cdot \mv{h}_2 + \mv{b}_3 \Big) \\            
            &= \sigma\Big( W_3 \cdot \sigma \big( W_2 \cdot \sigma (W_1\cdot \mv{h}_0 + \mv{b}_1) + \mv{b}_2 \big) + \mv{b}_3 \Big) \, .
\end{aligned}
\end{equation}
Let
\begin{equation} \label{eq:MLP-iterative-ftheta-j}
    f_{\theta_j}(\mv{z}) = \sigma(W_j \cdot \mv{z} + \mv{b}_j)
\end{equation}
and define   
\begin{equation} \label{eq:MLP-iterative-matrix-T}
    M_{MLP} =
    \begin{bmatrix}
        0                 & 0                 & 0                 & \dots  & 0  & 0 \\
        \Wg{f_{\theta_1}} & 0                 & 0                 & \dots  & 0  & 0\\
        0                 & \Wg{f_{\theta_2}} & 0                 & \dots  & 0  & 0\\
        0                 & 0                 & \Wg{f_{\theta_3}} & \dots  & 0  & 0 \\
        \vdots            & \vdots            &\vdots             & \ddots & \vdots & \vdots \\
        0                 & 0                 & 0                 & 0      & \Wg{f_{\theta_{T-1}}} & 0
    \end{bmatrix}
\end{equation}
where $f_{\theta_1}: \mathbb{R}^{|\mv{x}|} \mapsto \mathbb{R}^{|\mv{h}_1|}$ and
$\Wg{f_{\theta_j}}: \mathbb{R}^{ |\mv{h}_{j-1}|} \mapsto \mathbb{R}^{|\mv{h}_j|},\, j=2,...,T$, 
and initial conditions
\begin{equation} \label{eq:MLP-iterative-initial-condition-T}
     \begin{bmatrix}
        \mv{h}_0 \\
        0 \\
        0 \\
        \vdots \\
        0 \\
    \end{bmatrix} \, .
\end{equation}
This is an example of a \emph{sparse} block linear function, where the majority of the entries are the zero-function and only a few entries are non-zero.  In particular, the non-zero entries are the functions $f_{\theta_1}, f_{\theta_2}, \dots, f_{\theta_{T-1}}$ are assumed to be nonlinear functions.  The function $M_{MLP}$ is an \emph{autonomous}, \emph{nonlinear} map. 
Now define
\begin{equation} \label{eq:MLP-iterative-matrix-3}
    M_{MLP3} =
    \begin{bmatrix}
        0                 & 0                 & 0                 & 0 \\
        \Wg{f_{\theta_1}} & 0                 & 0                 & 0 \\
        0                 & \Wg{f_{\theta_2}} & 0                 & 0 \\
        0                 & 0                 & \Wg{f_{\theta_3}} & 0 \\
    \end{bmatrix}.
\end{equation}
We show that three iteration of the map $M_{MLP3}$ is equivalent to a three layer MLP.

We apply the map $M_{MLP3}$ (written here as $M$ to simplify notation) to the initial vector 
\begin{equation} \label{eq:MLP-iterative-initial-condition-3}
     \begin{bmatrix}
        \mv{h}_0 \\
        0 \\
        0 \\
        0 \\
    \end{bmatrix}
\end{equation}
which has length $(|\mv{h}_0| + |\mv{h}_1| + |\mv{h}_2| + |\mv{h}_3|)$.
After the first iteration of the map $M$
\begin{equation} \label{eq:MLP-iteration-1}
    M \circ 
    \begin{bmatrix}
        \mv{h}_0 \\
        0 \\
        0 \\
        0 \\
    \end{bmatrix} 
     =  
    \begin{bmatrix}
        0                 & 0                 & 0                 & 0 \\
        \Wg{f_{\theta_1}} & 0                 & 0                 & 0 \\
        0                 & \Wg{f_{\theta_2}} & 0                 & 0 \\
        0                 & 0                 & \Wg{f_{\theta_3}} & 0 \\
    \end{bmatrix}
        \circ
    \begin{bmatrix}
        \mv{h}_0 \\
        0 \\
        0 \\
        0 \\
    \end{bmatrix} 
    = 
    \begin{bmatrix}
        0 \\
        \Wg{f_{\theta_1}}(\mv{h}_0) \\
        \Wg{f_{\theta_2}}(0) \\
        \Wg{f_{\theta_3}}(0) \\
    \end{bmatrix} \, .
\end{equation}
After the second iteration of map $M$
\begin{equation} \label{eq:MLP-iteration-2}
    M \circ 
    M \circ 
    \begin{bmatrix}
        \mv{h}_0 \\
        0 \\
        0 \\
        0 \\
    \end{bmatrix} 
    = 
    \begin{bmatrix}
        0                 & 0                 & 0                 & 0 \\
        \Wg{f_{\theta_1}} & 0                 & 0                 & 0 \\
        0                 & \Wg{f_{\theta_2}} & 0                 & 0 \\
        0                 & 0                 & \Wg{f_{\theta_3}} & 0 \\
    \end{bmatrix}
    \circ 
    \begin{bmatrix}
        0 \\
        \Wg{f_{\theta_1}}(\mv{h}_0) \\
        \Wg{f_{\theta_2}}(0) \\
        \Wg{f_{\theta_3}}(0) \\
    \end{bmatrix} 
    =
    \begin{bmatrix}
        0 \\
        \Wg{f_{\theta_1}}(0) \\
        \Wg{f_{\theta_2}}(\Wg{f_{\theta_1}}(\mv{h}_0)) \\
        \Wg{f_{\theta_3}}(\Wg{f_{\theta_2}}(0))  \\
    \end{bmatrix} 
\end{equation}
and after the third iteration of  map $M$ 
\begin{equation} \label{eq:MLP-iteration-3}
\begin{aligned}
    M \circ 
    M \circ 
    M \circ 
    \begin{bmatrix}
        \mv{h}_0 \\
        0 \\
        0 \\
        0 \\
    \end{bmatrix} 
    &= 
    \begin{bmatrix}
        0                 & 0                 & 0                 & 0 \\
        \Wg{f_{\theta_1}} & 0                 & 0                 & 0 \\
        0                 & \Wg{f_{\theta_2}} & 0                 & 0 \\
        0                 & 0                 & \Wg{f_{\theta_3}} & 0 \\
    \end{bmatrix}
    \circ 
    \begin{bmatrix}
        0 \\
        \Wg{f_{\theta_1}}(0) \\
        \Wg{f_{\theta_2}}(\Wg{f_{\theta_1}}(\mv{h}_0)) \\
        \Wg{f_{\theta_3}}(\Wg{f_{\theta_2}}(0))  \\
    \end{bmatrix}   \\ 
    &= 
    \begin{bmatrix}
        0 \\
        \Wg{f_{\theta_1}}(0) \\
        \Wg{f_{\theta_2}}(\Wg{f_{\theta_1}}(0)) \\
        \Wg{f_{\theta_3}}(\Wg{f_{\theta_2}}(\Wg{f_{\theta_1}}(\mv{h}_0)))  \\
    \end{bmatrix} \, .  
\end{aligned}
\end{equation}

If we now interpret the last entry in the vector as the output of the map we observe that it is \emph{identical} to the output of a three layer MLP with weights $\Wg{f_{\theta_1}}$, $\Wg{f_{\theta_2}}$, and $\Wg{f_{\theta_3}}$ and input $\mv{h}_0$ as shown in \eqref{eq:MLP-traditional-compact-h3}.  Note, this is not a special property of this particular MLP, but instead a universal property of MLPs.  

\subsection{MLPs with additional iterations} \label{sec:MLP-additional-iterations}
\subsubsection{Finite impulse response} \label{sec:MLP-finite-impulse-response}
In \S \ref{sec:MLP-block-iterative} it is important to recognize that \eqref{eq:MLP-iterative-matrix-T} starting with initial conditions \eqref{eq:MLP-iterative-initial-condition-T} must be iterated $T$ times to achieve the equivalent outcome as \eqref{eq:MLP-traditional-hT}.  Further iterations lead to interesting results as shown below.

Applying the map $M$ one additional time reveals 
\begin{equation} \label{eq:MLP-iteration-4}
\begin{aligned}
    M \circ 
    M \circ 
    M \circ 
    M \circ 
    \begin{bmatrix}
        \mv{h}_0 \\
        0 \\
        0 \\
        0 \\
    \end{bmatrix} 
    &= 
    \begin{bmatrix}
        0                 & 0                 & 0                 & 0 \\
        \Wg{f_{\theta_1}} & 0                 & 0                 & 0 \\
        0                 & \Wg{f_{\theta_2}} & 0                 & 0 \\
        0                 & 0                 & \Wg{f_{\theta_3}} & 0 \\
    \end{bmatrix}
    \circ 
    \begin{bmatrix}
        0 \\
        \Wg{f_{\theta_1}}(0) \\
        \Wg{f_{\theta_2}}(\Wg{f_{\theta_1}}(0)) \\
        \Wg{f_{\theta_3}}(\Wg{f_{\theta_2}}(\Wg{f_{\theta_1}}(\mv{h}_0))))  \\
    \end{bmatrix}   \\
    &= 
   \begin{bmatrix}
        0 \\
        \Wg{f_{\theta_1}}(0) \\
        \Wg{f_{\theta_2}}(\Wg{f_{\theta_1}}(0)) \\
        \Wg{f_{\theta_3}}(\Wg{f_{\theta_2}}(\Wg{f_{\theta_1}}(0)))  \\
    \end{bmatrix}  
\end{aligned}
\end{equation}
which is independent of the input $\mv{h}_0$.  Accordingly, it is important to note that $M\circ M \circ M$ is identical to the original three-layer MLP, but $M\circ M \circ M \circ M$ is a totally different operator.  In fact, $M\circ M \circ M \circ M$ is finite impulse in that it does not depend on the input data $\mv{h}_0$.

\subsubsection{Fixed point} \label{sec:MLP-fixed-point}
An additional application of $M$, while seemingly inane from the point of view of NNs, does have interesting implications from the point of view of dynamical systems.  In particular, one may observe that

\begin{equation} \label{eq:MLP-iteration-5}
\begin{aligned}
    M \circ 
    M \circ 
    M \circ 
    M \circ
    M \circ
    \begin{bmatrix}
        \mv{h}_0 \\
        0 \\
        0 \\
        0 \\
    \end{bmatrix}
    &=
    \begin{bmatrix}
        0                 & 0                 & 0                 & 0 \\
        \Wg{f_{\theta_1}} & 0                 & 0                 & 0 \\
        0                 & \Wg{f_{\theta_2}} & 0                 & 0 \\
        0                 & 0                 & \Wg{f_{\theta_3}} & 0 \\
    \end{bmatrix}
    \circ 
    \begin{bmatrix}
        0 \\
        \Wg{f_{\theta_1}}(0) \\
        \Wg{f_{\theta_2}}(\Wg{f_{\theta_1}}(0)) \\
        \Wg{f_{\theta_3}}(\Wg{f_{\theta_2}}(\Wg{f_{\theta_1}}(0)))  \\
    \end{bmatrix}    \\
    &=  
    \begin{bmatrix}
        0 \\
        \Wg{f_{\theta_1}}(0) \\
        \Wg{f_{\theta_2}}(\Wg{f_{\theta_1}}(0)) \\
        \Wg{f_{\theta_3}}(\Wg{f_{\theta_2}}(\Wg{f_{\theta_1}}(0)))  \\
    \end{bmatrix}    
\end{aligned}
\end{equation}
which in addition, shows that 
\begin{equation} \label{eq:MLP-iteration-fixed-point}
   \begin{bmatrix}
        0 \\
        \Wg{f_{\theta_1}}(0) \\
        \Wg{f_{\theta_2}}(\Wg{f_{\theta_1}}(0)) \\
        \Wg{f_{\theta_3}}(\Wg{f_{\theta_2}}(\Wg{f_{\theta_1}}(0)))  \\
    \end{bmatrix} 
\end{equation}
is a fixed point of the map $M$.  Such examples underscore the tight, and perhaps surprising, relationship between NNs and dynamical systems, and now the dynamical system point of view provides insights into the deep connection between NNs and dynamical systems.  

\subsection{Choice of the initial vector} \label{sec:MLP3-q}

We apply the map $M_{MLP3}$ (written here as $M$ to simplify notation) to the initial vector 
\begin{equation} \label{eq:MLP-iterative-initial-condition-3-q}
     \begin{bmatrix}
        \mv{h}_0 \\
        q_1 \\
        q_2 \\
        q_3 \\
    \end{bmatrix}
\end{equation}
where $q_1, q_2, q_3 \not= 0$, which has length $(|\mv{h}_0| + |\mv{h}_1| + |\mv{h}_2| + |\mv{h}_3|)$.
After the first iteration of the map $M$
\begin{equation} \label{eq:MLP-iteration-1q}
    M \circ 
    \begin{bmatrix}
        \mv{h}_0 \\
        q_1 \\
        q_2 \\
        q_3 \\
    \end{bmatrix} 
     =  
    \begin{bmatrix}
        0                 & 0                 & 0                 & 0 \\
        \Wg{f_{\theta_1}} & 0                 & 0                 & 0 \\
        0                 & \Wg{f_{\theta_2}} & 0                 & 0 \\
        0                 & 0                 & \Wg{f_{\theta_3}} & 0 \\
    \end{bmatrix}
        \circ
    \begin{bmatrix}
        \mv{h}_0 \\
        q_1 \\
        q_2 \\
        q_3 \\
    \end{bmatrix} 
    = 
    \begin{bmatrix}
        0 \\
        \Wg{f_{\theta_1}}(\mv{h}_0) \\
        \Wg{f_{\theta_2}}(q_1) \\
        \Wg{f_{\theta_3}}(q_2) \\
    \end{bmatrix} \, .
\end{equation}
After the second iteration of map $M$
\begin{equation} \label{eq:MLP-iteration-2q}
    M \circ 
    M \circ 
    \begin{bmatrix}
        \mv{h}_0 \\
        q_1 \\
        q_2 \\
        q_3 \\
    \end{bmatrix} 
    = 
    \begin{bmatrix}
        0                 & 0                 & 0                 & 0 \\
        \Wg{f_{\theta_1}} & 0                 & 0                 & 0 \\
        0                 & \Wg{f_{\theta_2}} & 0                 & 0 \\
        0                 & 0                 & \Wg{f_{\theta_3}} & 0 \\
    \end{bmatrix}
    \circ 
    \begin{bmatrix}
        0 \\
        \Wg{f_{\theta_1}}(\mv{h}_0) \\
        \Wg{f_{\theta_2}}(q_1) \\
        \Wg{f_{\theta_3}}(q_2) \\
    \end{bmatrix}
    =
    \begin{bmatrix}
        0 \\
        \Wg{f_{\theta_1}}(0) \\
        \Wg{f_{\theta_2}}(\Wg{f_{\theta_1}}(\mv{h}_0)) \\
        \Wg{f_{\theta_3}}(\Wg{f_{\theta_2}}(q_1))  \\
    \end{bmatrix} 
\end{equation}
and after the third iteration of  map $M$ 
\begin{equation} \label{eq:MLP-iteration-3q}
\begin{aligned}
    M \circ 
    M \circ 
    M \circ 
    \begin{bmatrix}
        \mv{h}_0 \\
        q_1 \\
        q_2 \\
        q_3 \\
    \end{bmatrix} 
    &= 
    \begin{bmatrix}
        0                 & 0                 & 0                 & 0 \\
        \Wg{f_{\theta_1}} & 0                 & 0                 & 0 \\
        0                 & \Wg{f_{\theta_2}} & 0                 & 0 \\
        0                 & 0                 & \Wg{f_{\theta_3}} & 0 \\
    \end{bmatrix}
    \circ 
    \begin{bmatrix}
        0 \\
        \Wg{f_{\theta_1}}(0) \\
        \Wg{f_{\theta_2}}(\Wg{f_{\theta_1}}(\mv{h}_0)) \\
        \Wg{f_{\theta_3}}(\Wg{f_{\theta_2}}(q_1))  \\
    \end{bmatrix}   \\ 
    &= 
    \begin{bmatrix}
        0 \\
        \Wg{f_{\theta_1}}(0) \\
        \Wg{f_{\theta_2}}(\Wg{f_{\theta_1}}(0)) \\
        \Wg{f_{\theta_3}}(\Wg{f_{\theta_2}}(\Wg{f_{\theta_1}}(\mv{h}_0)))  \\
    \end{bmatrix} \, . 
\end{aligned}
\end{equation}
The solution after three iterations is identical to that in \eqref{eq:MLP-iteration-3}. Clearly then, further iterations result in
\begin{equation} \label{eq:MLP-iteration-4q}
\begin{aligned}
    M \circ 
    M \circ 
    M \circ 
    M \circ 
    \begin{bmatrix}
        \mv{h}_0 \\
        q_1 \\
        q_2 \\
        q_3 \\
    \end{bmatrix} 
    &= 
    \begin{bmatrix}
        0                 & 0                 & 0                 & 0 \\
        \Wg{f_{\theta_1}} & 0                 & 0                 & 0 \\
        0                 & \Wg{f_{\theta_2}} & 0                 & 0 \\
        0                 & 0                 & \Wg{f_{\theta_3}} & 0 \\
    \end{bmatrix}
    \circ 
    \begin{bmatrix}
        0 \\
        \Wg{f_{\theta_1}}(0) \\
        \Wg{f_{\theta_2}}(\Wg{f_{\theta_1}}(0)) \\
        \Wg{f_{\theta_3}}(\Wg{f_{\theta_2}}(\Wg{f_{\theta_1}}(\mv{h}_0))))  \\
    \end{bmatrix} \\
    &= 
    \begin{bmatrix}
        0 \\
        \Wg{f_{\theta_1}}(0) \\
        \Wg{f_{\theta_2}}(\Wg{f_{\theta_1}}(0)) \\
        \Wg{f_{\theta_3}}(\Wg{f_{\theta_2}}(\Wg{f_{\theta_1}}(0)))  \\
    \end{bmatrix}   
\end{aligned}
\end{equation}
and
\begin{equation} \label{eq:MLP-iteration-5q}
\begin{aligned} 
    M \circ 
    M \circ 
    M \circ 
    M \circ 
    M \circ 
    \begin{bmatrix}
        \mv{h}_0 \\
        q_1 \\
        q_2 \\
        q_3 \\
    \end{bmatrix} 
    = 
    \begin{bmatrix}
        0                 & 0                 & 0                 & 0 \\
        \Wg{f_{\theta_1}} & 0                 & 0                 & 0 \\
        0                 & \Wg{f_{\theta_2}} & 0                 & 0 \\
        0                 & 0                 & \Wg{f_{\theta_3}} & 0 \\
    \end{bmatrix}
    \circ 
    \begin{bmatrix}
        0 \\
        \Wg{f_{\theta_1}}(0) \\
        \Wg{f_{\theta_2}}(\Wg{f_{\theta_1}}(0)) \\
        \Wg{f_{\theta_3}}(\Wg{f_{\theta_2}}(\Wg{f_{\theta_1}}(0)))  \\
    \end{bmatrix}    
    = \\
    \begin{bmatrix}
        0 \\
        \Wg{f_{\theta_1}}(0) \\
        \Wg{f_{\theta_2}}(\Wg{f_{\theta_1}}(0)) \\
        \Wg{f_{\theta_3}}(\Wg{f_{\theta_2}}(\Wg{f_{\theta_1}}(0)))  \\
    \end{bmatrix} \, .
\end{aligned}
\end{equation}

Note, again, the independence of the final result to the initial input vector.

\section{RNNs and MLPs as dynamical systems} \label{sec:dynamical-systems}

When viewed as iterative maps (dynamical systems), RNNs and MLPs have {\bf identical} structures as shown in \eqref{eq:RNN-iteration-3} and \eqref{eq:MLP-iteration-3}, namely
\begin{equation} \label{eq:DS-basic-iteration}
    \begin{bmatrix}
    0            & 0            & \dots  & 0 & 0 \\
    f_1 & 0            & \dots  & 0 & 0 \\
    0            & f_2 & \dots  & 0 & 0 \\
    \vdots       &              & \ddots  &   & \vdots \\
    0            & 0            & \dots & f_T & 0 
    \end{bmatrix} 
    \circ
    \begin{bmatrix}
    0            & 0            & \dots  & 0 & 0 \\
    f_1 & 0            & \dots  & 0 & 0 \\
    0            & f_2 & \dots  & 0 & 0 \\
    \vdots       &              & \ddots  &   & \vdots \\
    0            & 0            & \dots & f_T & 0 
    \end{bmatrix} 
    \circ \hdots \circ
    \begin{bmatrix}
    0            & 0            & \dots  & 0 & 0 \\
    f_1 & 0            & \dots  & 0 & 0 \\
    0            & f_2 & \dots  & 0 & 0 \\
    \vdots       &              & \ddots  &   & \vdots \\
    0            & 0            & \dots & f_T & 0 
    \end{bmatrix} 
    \circ
    \begin{bmatrix}
        \mv{h}_0  \\
        0 \\
        0 \\
        \vdots \\
        0
    \end{bmatrix}.
\end{equation}
The difference lies in the parameter sets $\theta_{RNN}$ and $\theta_{MLP}$ and in the definitions of the functions $f_j$, specifically
\begin{equation} \label{eq:DS-RNN-vs-MLP}
\begin{aligned}
    &\hbox{RNN:} \qquad \theta_{RNN} = \{W_x, W_h, \mv{b} \} \;  \\
    &\hbox{MLP:} \qquad \theta_{MLP} = \{W_j, \mv{b}_j \}, \; j=1 \dots, T \\
&\hspace{-1.5cm}\hbox{and} \\
    &\hbox{RNN:} \qquad   f = f_\theta(\mv{x}_j,\mv{h}_{j-1}) = \sigma(W_x \cdot \mv{x}_j + W_h \cdot\mv{h}_{j-1} + \mv{b}), \quad j=1, \dots, T \\
    &\hbox{MLP:}  \qquad   f_j = f_{\theta_j}(\mv{h}_{j-1}) = \sigma(W_j \cdot \mv{h}_{j-1} + \mv{b}_j),  \quad j=1, \dots, T \\
\end{aligned}
\end{equation}
where $\mv{x}_j$ is the forcing function in the case of the RNN and, similarily, $\mv{h}_0$ is the initialization data of both the MLP and RNN.  Thus, we observe \emph{both} RNNs and MLPs may be written in the form  of \eqref{eq:DS-basic-iteration} by merely noting that for RNNs we have $\forall i,j \in 1,\cdots,T, \; f_i = f_j = f$.

\bigskip
\noindent\emph{The iterative map in \eqref{eq:DS-basic-iteration} plays two key roles in our work:}

\begin{itemize}
\item It provides a unifying perspective for understanding the relationship between MLPs and RNNs.  In fact, they are both dynamical systems and can be explored from that 
point of view. 
\item Equation \eqref{eq:DS-basic-iteration} also underscores the fact that a dynamical systems perspective inspires one to define and explore families of functions which are more general then those normally considered as MLPs and RNNs.  In fact, each of the $0$ blocks in \eqref{eq:DS-basic-iteration}, when trained and allowed to be non-zero, leads to a new family of functions with interesting theoretical and practical properties.
\end{itemize}

\noindent Exploring these two aspects of \eqref{eq:DS-basic-iteration} will comprise the remainder of this text.


\section{Modified MLP as an infinite impulse response map} \label{sec:nearby-infinite-impulse}

\subsection{Map with a non-zero diagonal entry} \label{sec:Minfty-q-0}

It is interesting to note, and will form the foundation of much future work, that seeming minor modifications to systems such as \eqref{eq:MLP-iteration-3} can lead to important differences in the resulting dynamical systems.  For example, adding an identity matrix to the top left corner of the map $M$ results in an infinite impulse response map. 
Define \footnote{Because of the $I$ on the diagonal of $M_\infty$ it is not a sequential operator.  However, we maintain the name Sequential2D for its implementation to engender the intuition that Sequential2D is a container for (non-linear) functions just like the more common Sequential container.}
\begin{equation} \label{eq:Minfty}
    M_{\infty} =
    \begin{bmatrix}
        \Wy{I}            & 0                 & 0                 & 0 \\
        \Wg{f_{\theta_1}} & 0                 & 0                 & 0 \\
        0                 & \Wg{f_{\theta_2}} & 0                 & 0 \\
        0                 & 0                 & \Wg{f_{\theta_3}} & 0 \\
    \end{bmatrix} \, .
\end{equation}
When started at the state 
\begin{equation} \label{eq:Minfty-ics}
    \begin{bmatrix}
        \mv{h}_0 \\
        0   \\
        0   \\
        0   \\
    \end{bmatrix}     
\end{equation}
after the first iteration of $M_\infty$
\begin{equation} \label{eq:Minfty-iteration-1}
    M_{\infty} \circ 
    \begin{bmatrix}
        \mv{h}_0 \\
        0   \\
        0   \\
        0   \\
    \end{bmatrix} 
     =  
    \begin{bmatrix}
        \Wy{I}            & 0                 & 0                 & 0 \\
        \Wg{f_{\theta_1}} & 0                 & 0                 & 0 \\
        0                 & \Wg{f_{\theta_2}} & 0                 & 0 \\
        0                 & 0                 & \Wg{f_{\theta_3}} & 0 \\
    \end{bmatrix}
        \circ
    \begin{bmatrix}
        \mv{h}_0 \\
        0   \\
        0   \\
        0   \\
    \end{bmatrix} 
    = 
    \begin{bmatrix}
        \mv{h}_0 \\
        \Wg{f_{\theta_1}}(\mv{h}_0) \\
        \Wg{f_{\theta_2}}(0) \\
        \Wg{f_{\theta_3}}(0) \\
    \end{bmatrix} \, .
\end{equation}
After the second iteration of map $M_{\infty}$ 
\begin{equation} \label{eq:Minfty-iteration-2}
    M_{\infty} \circ 
    M_{\infty} \circ 
    \begin{bmatrix}
        \mv{h}_0 \\
        0   \\
        0   \\
        0   \\
    \end{bmatrix} 
    = 
    \begin{bmatrix}
        \Wy{I}            & 0                 & 0                 & 0 \\
        \Wg{f_{\theta_1}} & 0                 & 0                 & 0 \\
        0                 & \Wg{f_{\theta_2}} & 0                 & 0 \\
        0                 & 0                 & \Wg{f_{\theta_3}} & 0 \\
    \end{bmatrix}
    \circ 
    \begin{bmatrix}
        \mv{h}_0 \\
        \Wg{f_{\theta_1}}(\mv{h}_0) \\
        \Wg{f_{\theta_2}}(0) \\
        \Wg{f_{\theta_3}}(0) \\
    \end{bmatrix} 
    =
    \begin{bmatrix}
        \mv{h}_0 \\
        \Wg{f_{\theta_1}}(\mv{h}_0) \\
        \Wg{f_{\theta_2}}(\Wg{f_{\theta_1}}(\mv{h}_0)) \\
        \Wg{f_{\theta_3}}(\Wg{f_{\theta_2}}(0)) \\
    \end{bmatrix} 
\end{equation}
and after the third iteration of  map $M_{\infty}$
\begin{equation} \label{eq:Minfty-iteration-3}
\begin{aligned}
    M_{\infty} \circ 
    M_{\infty} \circ 
    M_{\infty} \circ 
    \begin{bmatrix}
        \mv{h}_0 \\
        0 \\
        0 \\
        0 \\
    \end{bmatrix} 
    &= 
    \begin{bmatrix}
        \Wy{I}            & 0                 & 0                 & 0 \\
        \Wg{f_{\theta_1}} & 0                 & 0                 & 0 \\
        0                 & \Wg{f_{\theta_2}} & 0                 & 0 \\
        0                 & 0                 & \Wg{f_{\theta_3}} & 0 \\
    \end{bmatrix}
    \circ 
    \begin{bmatrix}
        \mv{h}_0 \\
        \Wg{f_{\theta_1}}(\mv{h}_0) \\
        \Wg{f_{\theta_2}}(\Wg{f_{\theta_1}}(\mv{h}_0)) \\
        \Wg{f_{\theta_3}}(\Wg{f_{\theta_2}}(0)) \\
    \end{bmatrix} \\ 
    &= 
    \begin{bmatrix}
        \mv{h}_0 \\
        \Wg{f_{\theta_1}}(\mv{h}_0) \\
        \Wg{f_{\theta_2}}(\Wg{f_{\theta_1}}(\mv{h}_0)) \\
        \Wg{f_{\theta_3}}(\Wg{f_{\theta_2}}(\Wg{f_{\theta_1}}(\mv{h}_0)))  \\
    \end{bmatrix}.  
\end{aligned}
\end{equation}
We observe that the last entry is exactly the same (vector) value as the corresponding MLP with weights $\Wg{f_{\theta_3}}$, $\Wg{f_{\theta_2}}$, and $\Wg{f_{\theta_1}}$ and input $\mv{h}_0$. Further iterations give
\begin{equation} \label{eq:Minfty-iteration-4}
\begin{aligned}
    M_{\infty} \circ
    M_{\infty} \circ 
    M_{\infty} \circ 
    M_{\infty} \circ 
    \begin{bmatrix}
        \mv{h}_0 \\
        0   \\
        0   \\
        0   \\
    \end{bmatrix} 
    &= 
    \begin{bmatrix}
        \Wy{I}            & 0                 & 0                 & 0 \\
        \Wg{f_{\theta_1}} & 0                 & 0                 & 0 \\
        0                 & \Wg{f_{\theta_2}} & 0                 & 0 \\
        0                 & 0                 & \Wg{f_{\theta_3}} & 0 \\
    \end{bmatrix}
    \circ 
    \begin{bmatrix}
        \mv{h}_0 \\
        \Wg{f_{\theta_1}}(\mv{h}_0) \\
        \Wg{f_{\theta_2}}(\Wg{f_{\theta_1}}(\mv{h}_0)) \\
        \Wg{f_{\theta_3}}(\Wg{f_{\theta_2}}(\Wg{f_{\theta_1}}(\mv{h}_0)))  \\
    \end{bmatrix} \\
    &=
    \begin{bmatrix}
        \mv{h_0} \\
        \Wg{f_{\theta_1}}(\mv{h}_0)) \\
        \Wg{f_{\theta_2}}(\Wg{f_{\theta_1}}(\mv{h}_0)) \\
        \Wg{f_{\theta_3}}(\Wg{f_{\theta_2}}(\Wg{f_{\theta_1}}(\mv{h}_0)))  \\
    \end{bmatrix}
\end{aligned}
\end{equation}
which is a fixed point that is dependent on the input $\mv{h}_0$, hence $M_{\infty}$ is a dynamical system with infinite impulse response. 

Summarizing, the maps defined by \eqref{eq:MLP-iterative-matrix-3} and \eqref{eq:Minfty} both converge to a fixed point.  The fixed point defined by \eqref{eq:MLP-iterative-matrix-3} is independent of the input and therefore is a finite impulse response map.  The fixed point defined by \eqref{eq:Minfty} is dependent of the input and therefore is a infinite impulse response map.


\subsection{Choice of the initial vector}  \label{sec:Minfty-q}

In \S \ref{sec:Minfty-q} we observed that the result of the iterative map only depends on $h_0$ and is invariant to the other entries of the initial vector, as long as those entries are $0$.  However, if, as we claim, the iterative map is equivalent to an MLP, then the iterative map should be invariant to the values in the initial vector, other than $h_0$ \emph{regardless of their value}.  Accordingly, in this section we consider the same iterative map as in \S \ref{sec:Minfty-q}, but with a different initial vector.  In particular, we consider the initial vector

\begin{equation} \label{eq:nearby-maps-Minfty-q}
    \begin{bmatrix}
        \mv{h}_0 \\
        q_1 \\
        q_2 \\
        q_3 \\
    \end{bmatrix}   
\end{equation}
after the first iteration of map $M_\infty$
\begin{equation} \label{eq:Minfty-iteration-1q}
    M_{\infty} \circ 
    \begin{bmatrix}
        \mv{h}_0 \\
        q_1 \\
        q_2 \\
        q_3 \\
    \end{bmatrix} 
     =  
    \begin{bmatrix}
        \Wy{I}            & 0                 & 0                 & 0 \\
        \Wg{f_{\theta_1}} & 0                 & 0                 & 0 \\
        0                 & \Wg{f_{\theta_2}} & 0                 & 0 \\
        0                 & 0                 & \Wg{f_{\theta_3}} & 0 \\
    \end{bmatrix}
        \circ
    \begin{bmatrix}
        \mv{h}_0 \\
        q_1 \\
        q_2 \\
        q_3 \\
    \end{bmatrix} 
    = 
    \begin{bmatrix}
        \mv{h}_0 \\
        \Wg{f_{\theta_1}}(\mv{h}_0) \\
        \Wg{f_{\theta_2}}(q_1) \\
        \Wg{f_{\theta_3}}(q_2) \\
    \end{bmatrix} \, .
\end{equation}
After the second iteration of map $M_\infty$
\begin{equation} \label{eq:Minfty-iteration-2q}
    M_{\infty} \circ 
    M_{\infty} \circ 
    \begin{bmatrix}
        \mv{h}_0 \\
        q_1 \\
        q_2 \\
        q_3 \\
    \end{bmatrix} 
    = 
    \begin{bmatrix}
        \Wy{I}            & 0                 & 0                 & 0 \\
        \Wg{f_{\theta_1}} & 0                 & 0                 & 0 \\
        0                 & \Wg{f_{\theta_2}} & 0                 & 0 \\
        0                 & 0                 & \Wg{f_{\theta_3}} & 0 \\
    \end{bmatrix}
    \circ 
    \begin{bmatrix}
        \mv{h}_0 \\
        \Wg{f_{\theta_1}}(\mv{h}_0) \\
        \Wg{f_{\theta_2}}(q_1) \\
        \Wg{f_{\theta_3}}(q_2) \\
    \end{bmatrix}
    =
    \begin{bmatrix}
        \mv{h}_0 \\
        \Wg{f_{\theta_1}}(\mv{h_0}) \\
        \Wg{f_{\theta_2}}(\Wg{f_{\theta_1}}(\mv{h}_0)) \\
        \Wg{f_{\theta_3}}(\Wg{f_{\theta_2}}(q_1))  \\
    \end{bmatrix} 
\end{equation}
and after the third iteration of  map $M$ 
\begin{equation} \label{eq:Minfty-iteration-3q}
\begin{aligned} 
    M_{\infty} \circ 
    M_{\infty} \circ 
    M_{\infty} \circ 
    \begin{bmatrix}
        \mv{h}_0 \\
        q_1 \\
        q_2 \\
        q_3 \\
    \end{bmatrix} 
    &= 
    \begin{bmatrix}
        \Wy{I}            & 0                 & 0                 & 0 \\
        \Wg{f_{\theta_1}} & 0                 & 0                 & 0 \\
        0                 & \Wg{f_{\theta_2}} & 0                 & 0 \\
        0                 & 0                 & \Wg{f_{\theta_3}} & 0 \\
    \end{bmatrix}
    \circ 
    \begin{bmatrix}
        \mv{h}_0 \\
        \Wg{f_{\theta_1}}(\mv{h_0}) \\
        \Wg{f_{\theta_2}}(\Wg{f_{\theta_1}}(\mv{h}_0)) \\
        \Wg{f_{\theta_3}}(\Wg{f_{\theta_2}}(q_1))  \\
    \end{bmatrix} \\
    &=
    \begin{bmatrix}
        \mv{h_0} \\
        \Wg{f_{\theta_1}}(\mv{h_0}) \\
        \Wg{f_{\theta_2}}(\Wg{f_{\theta_1}}(\mv{h_0})) \\
        \Wg{f_{\theta_3}}(\Wg{f_{\theta_2}}(\Wg{f_{\theta_1}}(\mv{h}_0)))  \\
    \end{bmatrix}   
\end{aligned}
\end{equation}
identical to \eqref{eq:Minfty-iteration-3} and independent of $q_1, q_2, q_3$. Clearly then 
\begin{equation} \label{eq:Minfty-iteration-4q}
\begin{aligned} 
    M_{\infty} \circ
    M_{\infty} \circ 
    M_{\infty} \circ 
    M_{\infty} \circ 
    \begin{bmatrix}
        \mv{h}_0 \\
        q_1 \\
        q_2 \\
        q_3 \\
    \end{bmatrix} 
    &= 
    \begin{bmatrix}
        \Wy{I}            & 0                 & 0                 & 0 \\
        \Wg{f_{\theta_1}} & 0                 & 0                 & 0 \\
        0                 & \Wg{f_{\theta_2}} & 0                 & 0 \\
        0                 & 0                 & \Wg{f_{\theta_3}} & 0 \\
    \end{bmatrix}
    \circ
    \begin{bmatrix}
        \mv{h_0} \\
        \Wg{f_{\theta_1}}(\mv{h_0}) \\
        \Wg{f_{\theta_2}}(\Wg{f_{\theta_1}}(\mv{h_0})) \\
        \Wg{f_{\theta_3}}(\Wg{f_{\theta_2}}(\Wg{f_{\theta_1}}(\mv{h}_0)))  \\
    \end{bmatrix} \\
    &= 
    \begin{bmatrix}
        \mv{h}_0 \\
        \Wg{f_{\theta_1}}(\mv{h}_0) \\
        \Wg{f_{\theta_2}}(\Wg{f_{\theta_1}}(\mv{h}_0)) \\
        \Wg{f_{\theta_3}}(\Wg{f_{\theta_2}}(\Wg{f_{\theta_1}}(\mv{h}_0)))  \\
    \end{bmatrix}
\end{aligned}
\end{equation}
which is a fixed point that is dependent on the input $\mv{h}_0$ but independent of $q_1, q_2, q_3$.  


\subsection{Choice of the initial vector with below diagonal blocks}  
\label{sec:Minfty-q-below}
In this section we stray from the stand MLP models, and consider a more complicated block non-linear function. The dynamical system enables us to consider structures that are not so evident from the traditional perspective. In particular, 
we begin with the same initial vector
\begin{equation} \label{eq:nearby-maps-Minfty-q-below}
    \begin{bmatrix}
        \mv{h}_0 \\
        q_1 \\
        q_2 \\
        q_3 \\
    \end{bmatrix}   
\end{equation}
but consider the block non-linear function
\begin{equation} \label{eq:Minfty-iteration-1q-below}
    M_\text{skip}=
    \begin{bmatrix}
        \Wy{I}            & 0                 & 0                 & 0 \\
        \Wg{f_{\theta_1}} & 0                 & 0                 & 0 \\
        0                 & \Wg{f_{\theta_2}} & 0                 & 0 \\
        \Wb{S}            & 0                 & \Wg{f_{\theta_3}} & 0 \\
    \end{bmatrix}.
\end{equation}
Such a block non-linear function can be thought of as an MLP with a skip connection \cite{huang2017densely,he2016deep} from the input to the last layer. 

In this case, after the first iteration of map $M_\text{skip}$
\begin{equation} \label{eq:Minfty-iteration-1q-below-a}
    M_\text{skip} \circ 
    \begin{bmatrix}
        \mv{h}_0 \\
        q_1 \\
        q_2 \\
        q_3 \\
    \end{bmatrix} 
     =  
    \begin{bmatrix}
        \Wy{I}            & 0                 & 0                 & 0 \\
        \Wg{f_{\theta_1}} & 0                 & 0                 & 0 \\
        0                 & \Wg{f_{\theta_2}} & 0                 & 0 \\
        \Wb{S}            & 0                 & \Wg{f_{\theta_3}} & 0 \\
    \end{bmatrix}
        \circ
    \begin{bmatrix}
        \mv{h}_0 \\
        q_1 \\
        q_2 \\
        q_3 \\
    \end{bmatrix} 
    = 
    \begin{bmatrix}
        \mv{h}_0 \\
        \Wg{f_{\theta_1}}(\mv{h}_0) \\
        \Wg{f_{\theta_2}}(q_1) \\
        \Wb{S}(\mv{h}_0)+\Wg{f_{\theta_3}}(q_2) \\
    \end{bmatrix} \, .
\end{equation}
After the second iteration of map $M_\text{skip}$
\begin{equation}
\begin{aligned} \label{eq:Minfty-iteration-2q-below}
    M_\text{skip} \circ 
    M_\text{skip} \circ 
    \begin{bmatrix}
        \mv{h}_0 \\
        q_1 \\
        q_2 \\
        q_3 \\
    \end{bmatrix} 
    &= 
    \begin{bmatrix}
        \Wy{I}            & 0                 & 0                 & 0 \\
        \Wg{f_{\theta_1}} & 0                 & 0                 & 0 \\
        0                 & \Wg{f_{\theta_2}} & 0                 & 0 \\
        \Wb{S}            & 0                 & \Wg{f_{\theta_3}} & 0 \\
    \end{bmatrix}
    \circ 
    \begin{bmatrix}
        \mv{h}_0 \\
        \Wg{f_{\theta_1}}(\mv{h}_0) \\
        \Wg{f_{\theta_2}}(q_1) \\
        \Wb{S}(\mv{h}_0)+\Wg{f_{\theta_3}}(q_2) \\
    \end{bmatrix} \\
    &=
    \begin{bmatrix}
        \mv{h}_0 \\
        \Wg{f_{\theta_1}}(\mv{h_0}) \\
        \Wg{f_{\theta_2}}(\Wg{f_{\theta_1}}(\mv{h}_0)) \\
        \Wb{S}(\mv{h}_0)+\Wg{f_{\theta_3}}(\Wg{f_{\theta_2}}(q_1))  \\
    \end{bmatrix} 
\end{aligned}
\end{equation}
and after the third iteration of  map $M_\text{skip}$ 
\begin{equation} \label{eq:Minfty-iteration-3q-below}
\begin{aligned} 
    M_\text{skip} \circ 
    M_\text{skip} \circ 
    M_\text{skip} \circ 
    \begin{bmatrix}
        \mv{h}_0 \\
        q_1 \\
        q_2 \\
        q_3 \\
    \end{bmatrix} 
    &= 
    \begin{bmatrix}
        \Wy{I}            & 0                 & 0                 & 0 \\
        \Wg{f_{\theta_1}} & 0                 & 0                 & 0 \\
        0                 & \Wg{f_{\theta_2}} & 0                 & 0 \\
        \Wb{S}            & 0                 & \Wg{f_{\theta_3}} & 0 \\
    \end{bmatrix}
    \circ 
    \begin{bmatrix}
        \mv{h}_0 \\
        \Wg{f_{\theta_1}}(\mv{h_0}) \\
        \Wg{f_{\theta_2}}(\Wg{f_{\theta_1}}(\mv{h}_0)) \\
        \Wb{S}(\mv{h}_0)+\Wg{f_{\theta_3}}(\Wg{f_{\theta_2}}(q_1))  \\
    \end{bmatrix} \\
    &= 
    \begin{bmatrix}
        \mv{h_0} \\
        \Wg{f_{\theta_1}}(\mv{h_0}) \\
        \Wg{f_{\theta_2}}(\Wg{f_{\theta_1}}(\mv{h_0})) \\
        \Wb{S}(\mv{h}_0)+\Wg{f_{\theta_3}}(\Wg{f_{\theta_2}}(\Wg{f_{\theta_1}}(\mv{h}_0)))  \\
    \end{bmatrix}   
\end{aligned}
\end{equation}
which is independent of $q_1, q_2, q_3$. Clearly then 
\begin{equation} \label{eq:Minfty-iteration-4q-below}
\begin{aligned} 
    M_\text{skip} \circ
    M_\text{skip} \circ 
    &M_\text{skip} \circ 
    M_\text{skip} \circ 
    \begin{bmatrix}
        \mv{h}_0 \\
        q_1 \\
        q_2 \\
        q_3 \\
    \end{bmatrix} \\
    &= 
    \begin{bmatrix}
        \Wy{I}            & 0                 & 0                 & 0 \\
        \Wg{f_{\theta_1}} & 0                 & 0                 & 0 \\
        0                 & \Wg{f_{\theta_2}} & 0                 & 0 \\
        \Wb{S}            & 0                 & \Wg{f_{\theta_3}} & 0 \\
    \end{bmatrix}
    \circ
    \begin{bmatrix}
        \mv{h_0} \\
        \Wg{f_{\theta_1}}(\mv{h_0}) \\
        \Wg{f_{\theta_2}}(\Wg{f_{\theta_1}}(\mv{h_0})) \\
        \Wb{S}(\mv{h}_0)+\Wg{f_{\theta_3}}(\Wg{f_{\theta_2}}(\Wg{f_{\theta_1}}(\mv{h}_0)))  \\
    \end{bmatrix} \\
    &= 
    \begin{bmatrix}
        \mv{h}_0 \\
        \Wg{f_{\theta_1}}(\mv{h}_0) \\
        \Wg{f_{\theta_2}}(\Wg{f_{\theta_1}}(\mv{h}_0)) \\
        \Wb{S}(\mv{h}_0)+\Wg{f_{\theta_3}}(\Wg{f_{\theta_2}}(\Wg{f_{\theta_1}}(\mv{h}_0)))  \\
    \end{bmatrix} \, .
\end{aligned}
\end{equation}
Again, we see that the fixed point is dependent on the input $\mv{h}_0$ but independent of $q_1, q_2, q_3$. Of course,
the fixed point is different from the fixed point in \S \ref{sec:Minfty-q-0} and \S \ref{sec:Minfty-q}, because of the addition of the skip connections.
However, we have preserved the property that the fixed point is independent of the values of $q_1, q_2, q_3$ which is required by standard MLPs with skip connections \cite{huang2017densely,he2016deep}. 


\subsection{Choice of the initial vector with above diagonal blocks}  

\label{sec:Minfty-q-above}
Now, we consider a more complicated block non-linear function.  We know that RNNs are recurrent \cite{Goodfellow-et-al-2016} and, in many interesting cases, infinite-impulse \cite{miljanovic2012comparative}.  Accordingly, their final output normally depends on the value of their initial state.  Exploiting the dynamical system perspective again, it is now easy to represent systems that are no longer simple feed forward networks.  In this section we demonstrate such a dependence in a block non-linear function.

Similar to Section \ref{sec:Minfty-q} in this section we consider the same initial vector
\begin{equation} \label{eq:nearby-maps-Minfty-q-above}
    \begin{bmatrix}
        \mv{h}_0 \\
        q_1 \\
        q_2 \\
        q_3 \\
    \end{bmatrix}   
\end{equation}
but consider the block non-linear function
\begin{equation} \label{eq:Minfty-iteration-1q-above}
    M_\text{above}=
    \begin{bmatrix}
        \Wy{I}            & 0                 & 0                 & 0 \\
        \Wg{f_{\theta_1}} & 0                 & \Wr{S}            & 0 \\
        0                 & \Wg{f_{\theta_2}} & 0                 & 0 \\
        0                 & 0                 & \Wg{f_{\theta_3}} & 0 \\
    \end{bmatrix} \, .
\end{equation}
As the function $\Wr{S}$ is above the diagonal, this block non-linear function is infinite-impulse \cite{miljanovic2012comparative} unless $\Wr{S}$ is
specifically chosen to be zero, or some similar degenerate function. 

To demonstrate the dependence of \eqref{eq:Minfty-iteration-1q-above} on the initial vector \eqref{eq:nearby-maps-Minfty-q-above} we do the same calculation as before.  In particular, after the first iteration of map $M_\text{above}$
\begin{equation} \label{eq:Minfty-iteration-1q-above-a}
    M_\text{above} \circ 
    \begin{bmatrix}
        \mv{h}_0 \\
        q_1 \\
        q_2 \\
        q_3 \\
    \end{bmatrix} 
     =  
    \begin{bmatrix}
        \Wy{I}            & 0                 & 0                 & 0 \\
        \Wg{f_{\theta_1}} & 0                 & \Wr{S}            & 0 \\
        0                 & \Wg{f_{\theta_2}} & 0                 & 0 \\
        0                 & 0                 & \Wg{f_{\theta_3}} & 0 \\
    \end{bmatrix}
        \circ
    \begin{bmatrix}
        \mv{h}_0 \\
        q_1 \\
        q_2 \\
        q_3 \\
    \end{bmatrix} 
    = 
    \begin{bmatrix}
        \mv{h}_0 \\
        \Wg{f_{\theta_1}}(\mv{h}_0)  + \Wr{S}(q_2)\\
        \Wg{f_{\theta_2}}(q_1) \\
        \Wg{f_{\theta_3}}(q_2) \\
    \end{bmatrix} \, .
\end{equation}
After the second iteration of map $M_\text{above}$ 
\begin{equation} \label{eq:Minfty-iteration-2q-above}
\begin{aligned} 
    M_\text{above} \circ 
    M_\text{above} \circ 
    \begin{bmatrix}
        \mv{h}_0 \\
        q_1 \\
        q_2 \\
        q_3 \\
    \end{bmatrix} 
    &= 
    \begin{bmatrix}
        \Wy{I}            & 0                 & 0                 & 0 \\
        \Wg{f_{\theta_1}} & 0                 & \Wr{S}            & 0 \\
        0                 & \Wg{f_{\theta_2}} & 0                 & 0 \\
        0                 & 0                 & \Wg{f_{\theta_3}} & 0 \\
    \end{bmatrix}
    \circ 
    \begin{bmatrix}
        \mv{h}_0 \\
        \Wg{f_{\theta_1}}(\mv{h}_0)  + \Wr{S}(q_2) \\
        \Wg{f_{\theta_2}}(q_1) \\
        \Wg{f_{\theta_3}}(q_2) \\
    \end{bmatrix} \\
    &=
    \begin{bmatrix}
        \mv{h}_0 \\
        \Wg{f_{\theta_1}}(\mv{h_0}) + \Wr{S}(\Wg{f_{\theta_2}}(q_1))\\
        \Wg{f_{\theta_2}}(\Wg{f_{\theta_1}}(\mv{h}_0) + \Wr{S}(q_2) ) \\
        \Wg{f_{\theta_3}}(\Wg{f_{\theta_2}}(q_1))  \\
    \end{bmatrix} 
\end{aligned}
\end{equation}
and after the third iteration of  map $M_\text{above}$ 
\begin{equation} \label{eq:Minfty-iteration-3q-above}
\begin{aligned} 
    M_\text{above}\circ 
    M_\text{above} \circ 
    &M_\text{above} \circ 
    \begin{bmatrix}
        \mv{h}_0 \\
        q_1 \\
        q_2 \\
        q_3 \\
    \end{bmatrix} \\
    &= 
    \begin{bmatrix}
        \Wy{I}            & 0                 & 0                 & 0 \\
        \Wg{f_{\theta_1}} & 0                 & \Wr{S}            & 0 \\
        0                 & \Wg{f_{\theta_2}} & 0                 & 0 \\
        0                 & 0                 & \Wg{f_{\theta_3}} & 0 \\
    \end{bmatrix}
    \circ 
    \begin{bmatrix}
        \mv{h}_0 \\
        \Wg{f_{\theta_1}}(\mv{h_0}) + \Wr{S}(\Wg{f_{\theta_2}}(q_1)) \\
        \Wg{f_{\theta_2}}(\Wg{f_{\theta_1}}(\mv{h}_0) + \Wr{S}(q_2) ) \\
        \Wg{f_{\theta_3}}(\Wg{f_{\theta_2}}(q_1))  \\
    \end{bmatrix} \\
    &= 
    \begin{bmatrix}
        \mv{h_0} \\
        \Wg{f_{\theta_1}}(\mv{h_0}) + \Wr{S}(\Wg{f_{\theta_2}}(\Wg{f_{\theta_1}}(\mv{h}_0) + \Wr{S}(q_2)))\\
        \Wg{f_{\theta_2}}(\Wg{f_{\theta_1}}(\mv{h}_0) + \Wr{S}(\Wg{f_{\theta_2}}(q_1)) ) \\
        \Wg{f_{\theta_3}}(\Wg{f_{\theta_2}}(\Wg{f_{\theta_1}}(\mv{h}_0) + \Wr{S}(q_2)))  \\
    \end{bmatrix} \, .   
\end{aligned}
\end{equation}
Now, the fourth iteration of the map reveals that  
\begin{equation} \label{eq:Minfty-iteration-4q-above}
\begin{aligned} 
    M_\text{above}\circ
    &M_\text{above} \circ 
    M_\text{above} \circ 
    M_\text{above} \circ 
    \begin{bmatrix}
        \mv{h}_0 \\
        q_1 \\
        q_2 \\
        q_3 \\
    \end{bmatrix} \\
    &= 
    \begin{bmatrix}
        \Wy{I}            & 0                 & 0                 & 0 \\
        \Wg{f_{\theta_1}} & 0                 & \Wr{S}            & 0 \\
        0                 & \Wg{f_{\theta_2}} & 0                 & 0 \\
        0                 & 0                 & \Wg{f_{\theta_3}} & 0 \\
    \end{bmatrix}
    \circ 
    \begin{bmatrix}
        \mv{h_0} \\
        \Wg{f_{\theta_1}}(\mv{h_0}) + \Wr{S}(\Wg{f_{\theta_2}}(\Wg{f_{\theta_1}}(\mv{h}_0) + \Wr{S}(q_2)))\\
        \Wg{f_{\theta_2}}(\Wg{f_{\theta_1}}(\mv{h}_0) + \Wr{S}(\Wg{f_{\theta_2}}(q_1)) ) \\
        \Wg{f_{\theta_3}}(\Wg{f_{\theta_2}}(\Wg{f_{\theta_1}}(\mv{h}_0) + \Wr{S}(q_2)))   \\
    \end{bmatrix} \\
    &= 
    \begin{bmatrix}
        \mv{h}_0 \\
        \Wg{f_{\theta_1}}( \mv{h}_0 ) + \Wr{S}(\Wg{f_{\theta_2}}(\Wg{f_{\theta_1}}(\mv{h}_0) + \Wr{S}(\Wg{f_{\theta_2}}(q_1)) ))\\
        \Wg{f_{\theta_2}}( \Wg{f_{\theta_1}}(\mv{h_0}) + \Wr{S}(\Wg{f_{\theta_2}}(\Wg{f_{\theta_1}}(\mv{h}_0) + \Wr{S}(q_2))) ) \\
        \Wg{f_{\theta_3}}( \Wg{f_{\theta_2}}(\Wg{f_{\theta_1}}(\mv{h}_0) + \Wr{S}(\Wg{f_{\theta_2}}(q_1))) )  \\
    \end{bmatrix},
\end{aligned}
\end{equation}
which is \emph{not a fixed point} and is dependent on the input $\mv{h}_0$ as well as $q_1$ and $q_2$. Accordingly, this block non-linear function is not equivalent to an MLP, but instead displays the infinite impulse response of an RNN.   Of course, if we set $\Wr{S}$ to be zero, then the above calculation reduces to the calculation in \S \ref{sec:Minfty-q} and the fixed point is independent of $q_1, q_2, q_3$, and this demonstrates that MLPs and RNNs lay on a continuum of block non-linear functions.

\section{Implementation} \label{sec:implementation}

As discussed in \cite{pathak2023deep} there are many references in the literature that propose a block matrix structure for NNs.  Examples include, \cite{narang2017blocksparse} where the focus is reducing the computational complexity and memory requirements of RNNs while simultaneously maintaining accuracy.  Also, in \cite{narang2017blocksparse}, it is noted that block sparsity can lead to better performance on hardware architectures such as Graphical Processing Units that support block-sparse matrix multiplication.

Having derived various theoretical connections between MLPs and RNNs, we now take a brief detour into the implementation of such methods.  In particular, the block non-linear function from \eqref{eq:block-nonlinear-function} is closely related to the commonly used Sequential container that appears in both the deep learning libraries, PyTorch, \cite{pytorch2017automatic} and Keras, \cite{chollet2015keras}.  In fact, the function \eqref{eq:block-nonlinear-function} can be thought of as a two-dimensional generalization of the Sequential container, and we call this generalization \emph{Sequential2D}, which is described in more detail in \cite{pathak2023deep}.  A brief review of \emph{github.com}\footnote{Data gathered on 5-16-2023.} reveals the popularity of the Sequential architecture.  As discussed in \cite{pathak2023deep}, we noted 380,000 files leveraging Pytorch's or Keras's Sequential implementation.  Accordingly, the theoretical ideas described in this paper, and the practical implementation discussed in \cite{pathak2023deep}, promise to be able to impact many Neural Network (NN) applications. 

For the purposes of this paper, we merely observe that Sequential2D, \cite{pathak2023deep} refers to the 2D matrix of functions, as opposed to the 1D array of functions in a PyTorch or Keras Sequential container.  Specifically, closely following \cite{pathak2023deep}, we observe that Sequential2D is a PyTorch module that allows one to define a neural network by specifying a two dimensional array of modules.  Each individual module can be any type of layer or operation supported by PyTorch, \cite{pytorch2017automatic}.  Practically Sequential2D can be used to implement a variety of neural network models including MLPs, RNNs, and many other NNs.

\section{Numerical experiments} \label{sec:numerical-experiments}

Using the Sequential2D module as an implementation for the block non-linear function from Definition \ref{defn:block-nonlinear-function}, numerous experiments have been conducted seeking to examine performance of various models across a range of scenarios and conditions.  In particular, the interested reader can look to \cite{pathak2023deep, hershey2023deep, hershey2023exploring} for examples of the ideas in this paper being applied for practical machine learning problems.  We will not attempt to repeat the results from those works here, but merely endeavor to provide a sample of the surprising results that arise from treating MLPs and RNNs as dynamical systems.

\subsection{MNIST classification} 
\label{sec:numerical-experiment-MNIST-classification}

\begin{figure}[htbp]
     \centering
     \includegraphics[width=0.5\textwidth]{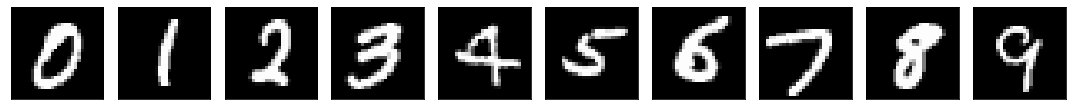}
     \caption{Examples of unmodified handwritten $28\times 28$ pixel handwritten images from the MNIST database.}
     \label{fig:mnist}
\end{figure}

In the following section we closely follow the presentation and results from \cite{hershey2023exploring}.  We study the classic MNIST, \cite{lecun2010mnist} (Modified National Institute of Standards and Technology) dataset, which is a common choice for machine learning exercises and features 70,000 labeled images often used for image recognition and classification problems. The content of each image is a hand written single digit integer ranging from 0-9 and a ground truth label reflecting the integer that was depicted.  Unmodified MNIST images can be seen in Fig. \ref{fig:mnist}.  In all experiments, the MNIST data set was divided into a training subset of 10,000 images, a validation subset of 1,000 images and a testing subset of 1,000 images with batch size of 64 trained for a duration of 100 epochs. 

Again, following \cite{hershey2023exploring}, several image transformations were applied with the objective of increasing the difficulty and creating more disparate comparisons among model structures.  Since model implementations were performed through PyTorch using the PyTorch Lightning wrapper (\cite{Falcon_PyTorch_Lightning_2019}), Torchvision (\cite{torchvision}) was a natural choice for transforms.  As a first step, images were resized (torchvision.transforms.Resize) from $28 \times 28$ to $50 \times 50$, increasing the total number of input features in $\mv{h}_0$  from 784 to 2,500.  Additionally, the features of each image are normalized (torchvision.transforms.Normalize) using the distribution defined as $N ~ (0.1307, 0.3081)$ in adherence with guidelines from the Torchvision documentation for the MNIST data set as shown in Figure \ref{fig:mnist}.

\begin{figure}[htbp] 
     \centering
     \begin{minipage}[b]{0.5\textwidth}
         \centering
         \includegraphics[width=\textwidth]{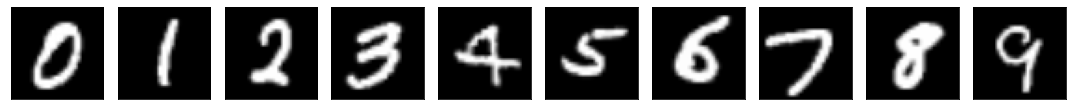}
         \label{fig:mnistbaseline}
     \end{minipage}
     \hfill
     \begin{minipage}[b]{0.5\textwidth}
         \centering
         \includegraphics[width=\textwidth]{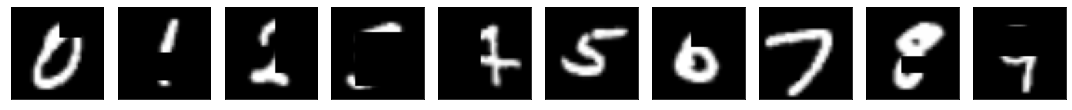}
         \label{fig:mnisterase}
     \end{minipage}
     \hfill
     \begin{minipage}[b]{0.5\textwidth}
         \centering
         \includegraphics[width=\textwidth]{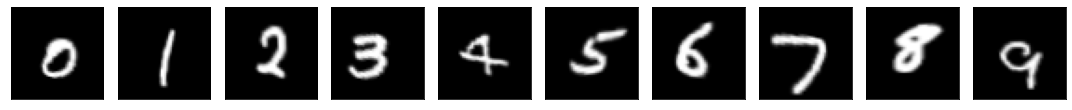}
         \label{fig:mnistperspective}
     \end{minipage}
     \hfill
     \begin{minipage}[b]{0.5\textwidth}
         \centering
         \includegraphics[width=\textwidth]{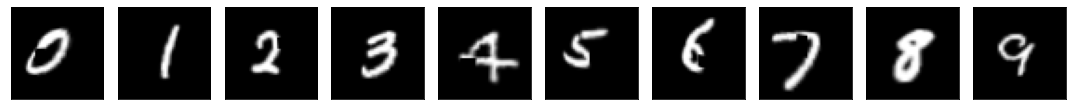}
         \label{fig:mnistboth}
     \end{minipage}
        \caption{The cumulative effects of transformations applied to samples from the MNIST data set. All images were first resized and then normalized. (row 1) Resized and normalized only; (row 2) with random erase as an intermediary step; (row 3) with random perspective as an intermediary step; (row 4) with both random erase and random perspective as intermediary steps.}
        \label{fig:basic-transformations}
\end{figure}

Once the image has been resized, each image in row 2 of Figure \ref{fig:basic-transformations} is transformed by randomly erasing (torchvision.transforms.RandomErasing) portions of the images prior to being normalized, creating a more challenging problem.  The amount of erasure is in the range (0.02, 0.05).  In row 3, each image is transformed after the normalization with a random perspective transformation (torchvision.transforms.RandomPerspective), using a distortion scale value of 0.5.  When applied in combination, both transforms from rows 2 and 3 result in the examples shown in row 4 of Figure \ref{fig:basic-transformations}, clearly an even more challenging problem.

\subsection{An MLP versus a dynamical system} \label{sec:numerical-experiment-model-structure}

For each run of the classification experiment for the MNIST data set, a base network was constructed to match the characteristics of a four-layer MLP with layers starting from an input dimension of 2,500 and with layer output dimensions of 500, 200, 100 and 10.  Following the notation of \S \ref{sec:MLP-block-iterative} this gives rise to a block non-linear function of the form

\begin{equation} \label{eq:MLP-example-iterative-matrix-4}
    M_{MLP4} =
    \begin{bmatrix}
        \Wy{I}            & 0                 & 0                 & 0                  & 0\\
        \Wg{f_{\theta_1}} & 0                 & 0                 & 0                  & 0\\
        0                 & \Wg{f_{\theta_2}} & 0                 & 0                  & 0\\
        0                 & 0                 & \Wg{f_{\theta_3}} & 0                  & 0\\
        0                 & 0                 & 0                 & \Wg{f_{\theta_4}}  & 0\\
    \end{bmatrix}.
\end{equation}

\noindent $f_{\theta_i}(\mv{z}) = \sigma(W_i \cdot \mv{z} + b_i)$ with
$I \in \mathbb{R}^{2500 \times 2500}$ being the identity function,
$\sigma$ being the standard ReLU activation function,
$\theta_1=\{W_1 \in \mathbb{R}^{500 \times 2500}, \mv{b}_1 \in \mathbb{R}^{500 \times 1}\}$,
$\theta_2=\{W_2 \in \mathbb{R}^{200 \times 500}, \mv{b}_2 \in \mathbb{R}^{200 \times 1}\}$,
$\theta_3=\{W_3 \in \mathbb{R}^{100 \times 200}, \mv{b}_3 \in \mathbb{R}^{100 \times 1}\}$, and
$\theta_4=\{W_4 \in \mathbb{R}^{10  \times 100}, \mv{b}_4 \in \mathbb{R}^{10 \times 1}\}$.
As demonstrated in \S \ref{sec:nearby-infinite-impulse}, such a block non-linear function is \emph{identical} to a standard MLP with the same layer sizes.

\begin{figure}[htbp] 
     \begin{minipage}{0.48\textwidth}
     \centering
     \begin{equation*} \label{eq:MLP-iterative-matrix-4}
        M_{MLP4} =
        \begin{bmatrix}
            \Wy{I}            & 0                 & 0                 & 0                  & 0\\
            \Wg{f_{\theta_1}} & 0                 & 0                 & 0                  & 0\\
            0                 & \Wg{f_{\theta_2}} & 0                 & 0                  & 0\\
            0                 & 0                 & \Wg{f_{\theta_3}} & 0                  & 0\\
            0                 & 0                 & 0                 & \Wg{f_{\theta_4}}  & 0\\
        \end{bmatrix}
     \end{equation*}  
     \end{minipage}   
     \hfill
     \begin{minipage}{0.40\textwidth}
     \centering
     \includegraphics[width=\textwidth]{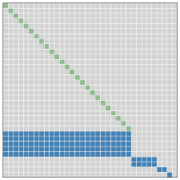}
     \end{minipage}
     \caption{Side-by-side comparison of an MLP represented as a dynamical system using a block non-linear function. The blue blocks represent trainable weights, the grey blocks represent non-trainable $0$ blocks, and the green blocks represent a $2500 \times 2500$ untrainable identity matrix. Note that this block non-linear function is of the same form at that discussed in \S \ref{sec:Minfty-q} with all of the properties discussed there.}
     \label{fig:iterativearchitectures}
\end{figure}

We emphasize the block structure inherent to the MLP in \eqref{eq:MLP-example-iterative-matrix-4} with a visualization in Figure \ref{fig:iterativearchitectures} in which the  block non-linear function is depicted ``to scale''.  The iteration matrix in \eqref{eq:MLP-example-iterative-matrix-4} is decomposed into smaller component matrices of size $100 \times 100$ except for the final row and column.  The last row is decomposed into component matrices of size $10 \times 100$.  The last column is decomposed in to component matrices of size  $100 \times 10$.  The matrix in the bottom right corner remained a square matrix of dimension $10 \times 10$.  The identity matrix in the top left hand corner is of size $2500 \times 2500$ or $25 \times 25, 100 \times 100$ blocks.  The blue blocks are trainable, the grey blocks are untrainable.  

\begin{figure}[htbp] 
    \centering
     \begin{minipage}{0.48\textwidth}
    \centering
    \begin{equation*} \label{eq:MLP-iterative-matrix-4-sparse}
        M_{MLP4} =
        \begin{bmatrix}
            \Wy{I} & 0      & 0      & 0      & 0      \\
            \Wg{S} & \Wy{S} & \Wr{S} & \Wr{S} & \Wr{S} \\
            \Wb{S} & \Wg{S} & \Wy{S} & \Wr{S} & \Wr{S} \\
            \Wb{S} & \Wb{S} & \Wg{S} & \Wy{S} & \Wr{S} \\
            \Wb{S} & \Wb{S} & \Wb{S} & \Wg{S} & \Wy{S} \\
        \end{bmatrix}
    \end{equation*}  
    \end{minipage}   
    \hfill
    \begin{minipage}{0.40\textwidth}
    \centering
    \includegraphics[width=\textwidth]{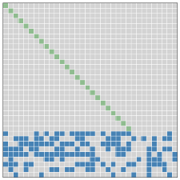}
    \end{minipage}
    \caption{A randomized dynamical system. While the positions of the trainable parameters are quite different when compared to Figure \ref{fig:iterativearchitectures}, many other properties are similar.  For example, the number of trainable parameters and the dimension of the input, hidden, and output spaces are all identical to \eqref{eq:MLP-example-iterative-matrix-4}. This block non-linear function combines the features of all the block non-linear functions in \S \ref{sec:Minfty-q}, \ref{sec:Minfty-q-below}, and, most importantly, \ref{sec:Minfty-q-above}. Despite these differences, we will demonstrate that the performance of this dynamical system is surprisingly indistinguishable from that of the MLP in Figure \ref{fig:iterativearchitectures}.}
    \label{fig:iterativearchitectures2}
\end{figure}

A question immediately arises when looking at Figure \ref{fig:iterativearchitectures}.  Is the placement of the blue trainable and grey untrainable blocks essential to the operation of the MLP?  For example, one can consider the ``randomized'' dynamical system shown in Figure \ref{fig:iterativearchitectures2}. 

Using the dynamical system architectures in Figure \ref{fig:iterativearchitectures} and Figure \ref{fig:iterativearchitectures2}, a traditional MNIST classification experiment was conducted, with the former dynamical system corresponding exactly to a standard MLP.  In contrast, the dynamical system in Figure \ref{fig:iterativearchitectures2} shares the same number of trainable parameters and dimensions of the input, hidden, and output spaces, but allocates those trainable parameters using a very different distribution.  Once the block nonlinear functions had been created, 10 training runs were performed on the data set using the traditional four layer MLP representation.  We then followed these 10 runs of the traditional MLP with 20 runs using randomized dynamical systems to explore the space of random configurations of trainable blocks and compare their performance. For both the architectures in Figures \ref{fig:iterativearchitectures} and \ref{fig:iterativearchitectures2} we use the same form of the initial vector $\mv{h}_0$, as in \eqref{eq:Minfty-ics} and shown below
\begin{equation} \label{eq:Minfty-iteration-5}
    \begin{aligned}    
    \begin{bmatrix}
        \mv{h}_0 \\
        0   \\
        0   \\
        0   \\
        0   \\
    \end{bmatrix}. 
\end{aligned}
\end{equation}
\noindent We emphasize the following points:
\begin{itemize}
\item For experiments that use the model in Figure \ref{fig:iterativearchitectures}, \emph{the values contained in the blue blocks are randomly initialized}.
\item For experiments that use the model in Figure \ref{fig:iterativearchitectures2}, \emph{both the placement of the blue values and the initial values contained in the blue blocks are randomized}.
\item For all experiments the initial vector in \eqref{eq:Minfty-iteration-5} \emph{is not random and they are fixed by the training data $h_0$ and the placement of the $0$s}.
\end{itemize}

\begin{figure}[htbp] 
     \centering
     \includegraphics[width=0.5\textwidth]{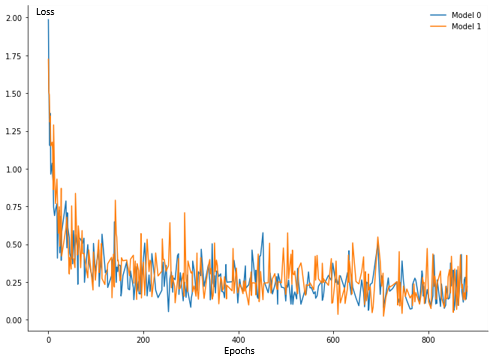}
     \caption{A comparison of training loss over 100 epochs between the layered architecture of Figure \ref{fig:iterativearchitectures} and the randomized architecture of Figure \ref{fig:iterativearchitectures2}. ``Model 0'' is the original MLP and ``Model 1'' is the randomized architecture. The performance curves are indistinguishable despite the dissimilar architectures.}
     \label{fig:iterativemlpcomparison}
\end{figure}

Following the process outlined above, experiments were conducted with results that are somewhat counter intuitive to prevailing theory regarding the principles and importance of network architecture.  As an illustration of these results, the two very different networks shown in Figure \ref{fig:iterativearchitectures} and Figure \ref{fig:iterativearchitectures2} were trained using the Adam optimizer from the PyTorch library with a learning rate of $1e-3$.  The resulting training profiles are presented in Figure \ref{fig:iterativemlpcomparison}.  Despite a stark difference in the distribution of the trainable parameters, both networks followed the same training profile with final testing accuracy of both models being virtually indistinguishable at 92.7\%.  These results, which clearly demonstrate an unexpected robustness, indicate other factors aside from layer based structure govern overall model performance.

\begin{figure}[htbp] 
     \centering
     \includegraphics[width=0.5\textwidth]{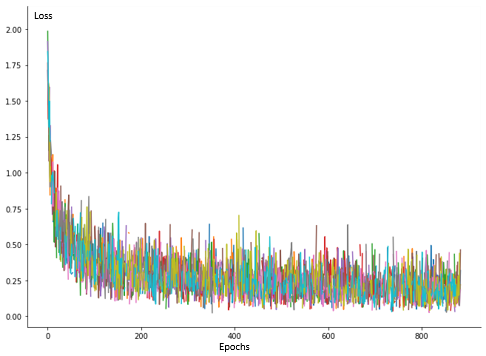}
     \caption{Performance curves demonstrating training loss over 100 epochs for 10 different random initializations of models with the layered architecture (Figure \ref{fig:iterativearchitectures}). As might be expected, the difference between the training of randomly initialized layered architectures is small. }  
     \label{fig:iterativemlpnonrandom}
\end{figure}

Figure \ref{fig:iterativemlpnonrandom} shows 10 
training curves for a four layer MLP.  Each of the training curves begins with a different, random initialization.  We see that all initializations result in similar training profiles.


\begin{figure}[htbp] 
     \centering
     \includegraphics[width=0.5\textwidth]{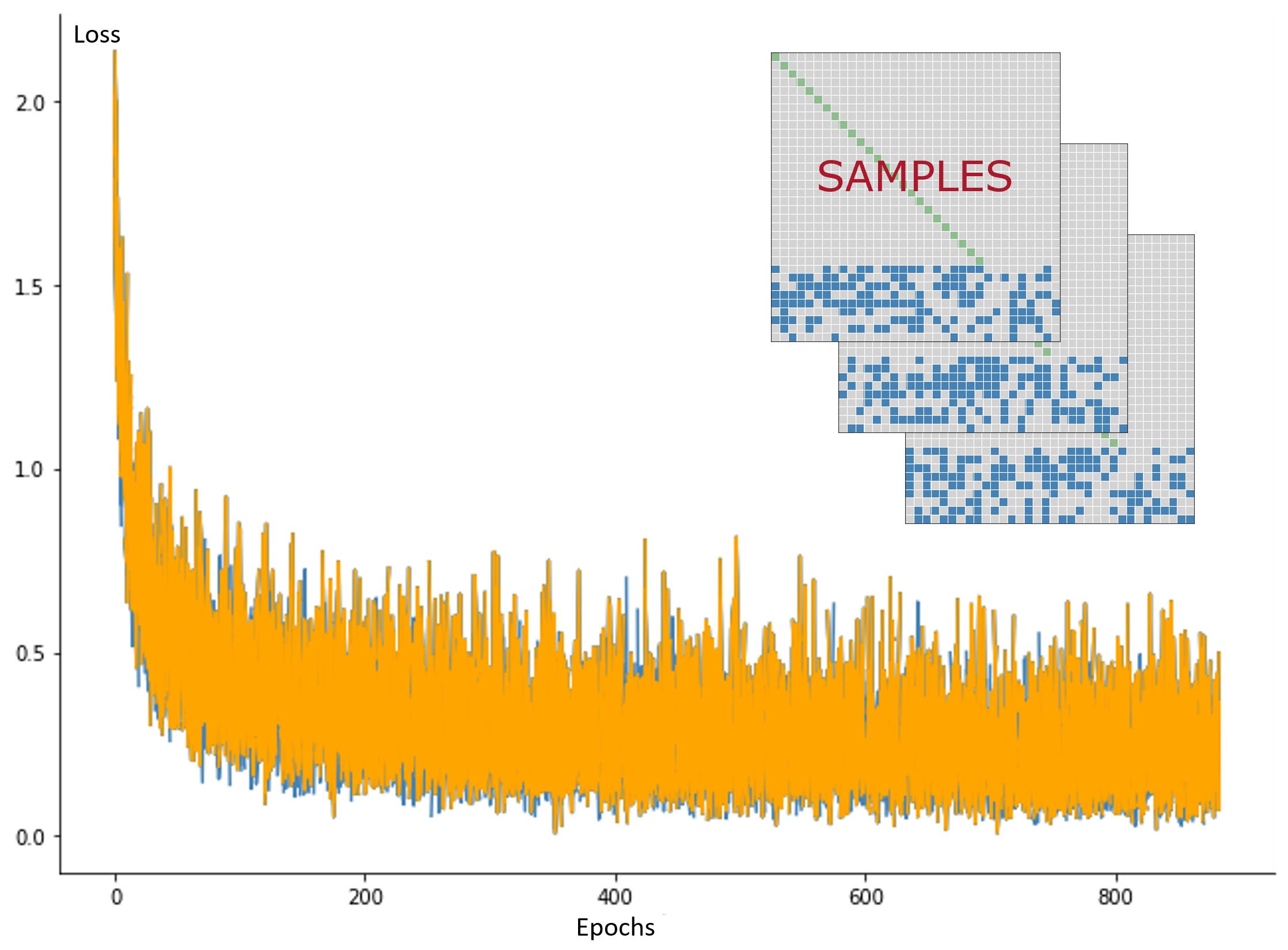}
     \caption{The training curves over 100 epochs with randomly initialized layered architectures are repeated from Figure \ref{fig:iterativemlpnonrandom} (blue). Training curves for models with randomly initialized randomized architectures (Figure \ref{fig:iterativearchitectures2}) over 100 epochs are overlaid in gold. The difference between the training of randomly initialized layered architectures is similar to the difference between the layered and randomized model architectures. \emph{This is quite suprising since the randomized architectures appear to be so different than the layered architectures.}}  
     \label{fig:iterativemlpcomparisons}
\end{figure}

In Figure \ref{fig:iterativemlpcomparisons} we plot the layered training loss curves from Figure \ref{fig:iterativemlpnonrandom} in blue before overlaying the training loss curves of models with randomized architectures (Figure \ref{fig:iterativearchitectures2}) in gold.  The results are compelling. When the 20 additional randomized training curves are overlaid they lie on top of one another.  With no similarity in structure, the models train with indistinguishable performance.  These results provide evidence that the concepts of layers and neatly defined weight matrices, as exist in Figure \ref{fig:iterativearchitectures}, seemingly provide no advantage over the randomized architectures in Figure \ref{fig:iterativearchitectures2}. \emph{Before seeing these results, one might have assumed that current MLPs represent the optimal network structure, but now this appears unlikely.}


\section{Summary}\label{sec:summary} 

The initial results in this paper demonstrate that despite major structural changes, randomized and layered architectures have little training variation.  This observation raises the valid question of whether the layered structure was ultimately irrelevant.  Specifically, if the arrangement of neatly demarcated layers is not a central feature of performance when normalized for the total number of trainable parameters, then the question becomes what factor or factors may be contributing to the performance of trained dynamical systems.  This question is explored further in \cite{hershey2023deep, hershey2023exploring}.

A dynamical system perspective offers a generalized context for exploring RNNs, MLPs and a vast array of other neural network architectures.  The results suggest layer wise architectural constraints have little impact on the training performance of at least some problems.  Other parameters, such as number of iterations (in the context of a dynamical system), sparsity, and other less commonly studied parameters may have a profound influence on model behavior.  We have only begun exploring this large space of machine learning models \cite{hershey2023deep, hershey2023exploring, pathak2023deep}, and we hope this paper will encourage others to study this domain. 

One important aspect of this work is that the number of iterations in the context of a dynamical system and the number of layers in a neural network are related but distinct concepts.  In particular a layered architecture forces a number of iterations of the dynamical system (see our discussion of fixed points).  However, a broader view of training dynamical systems decouples the architecture from the number of iterations.  The randomized dynamical system in Figure \ref{fig:iterativearchitectures2} was trained and tested exclusively over four iterations, but that is not required.  Exploring the relationship between different training paradigms given the freedom provided by the dynamical systems approach provides fertile ground for exploration.


The effect of noise in the data for both types of architectures remains an open issue.  A sensitivity analysis could provide insight into the effect of architecture on stability.  This is a natural question in a dynamical systems context with its concept of stable and unstable manifolds, but has no obvious analogue in standard approaches to the construction and analysis of neural networks.

Continuation and bifurcation analysis would seem to be another rich field of study for practical methods of training dynamical systems to solve real world problems.  For instance, we have already demonstrated that GPT2's performance can be substantially improved through a parameter continuation methodology inspired by the results of this paper and a common technique in dynamical systems \cite{pathak2023deep}.  Since a simple Sequential construction shown in \eqref{eq:MLP-iterative-matrix-T} is the basis of all MLPs, as well as many other neural networks, and can be embedded within the Sequential2D framework, it is reasonable to expect this new approach will provide improvements. 

A disadvantage of a dynamical systems approach where a fixed map is iterated many times to achieve some goal is in the training of such a map.  For example, the gradients that one computes using automatic differentiation are more complicated in this case.  The existence of advanced optimization libraries such as PyTorch and TensorFlow ameliorate many of the technical issues in computing such gradients efficiently.  However, this does not directly address such problems as gradient collapse and gradient explosion.  Yet, within the context of dynamical systems natural analogues exist for these characteristics.  The notions of stable and unstable manifolds correspond to gradient collapse and gradient explosion respectively.  Further, there exist many techniques in dynamical systems for addressing exactly these types of problems. 

As soon as one thinks about training dynamical systems to solve practical problems there are many interesting avenues for theoretical and performance analysis.  In particular Koopman operators are an active area of mathematical research that are precisely concerned with data oriented dynamical systems.  Our proposed approach allows us to immediately leverage this body of work. 

The stepwise nature of trained dynamical systems leads to an interesting trade-off between complicated dynamical systems that achieve some practical output in a fewer number of iterations versus simple dynamical systems that achieve the same output but only after a greater number of iterations.  While seemingly a bad trade-off, having a simple dynamical system that achieves a given result does provide advantages in analysis, explainability and safety.  In particular, the use of simple dynamical systems over many iterations would seem to address some of the goals of mechanistic interpretability in AI \cite{zhang2021survey,kastner2023explaining,shen2023large,michaud2024opening}.

\acks{This research was carried out using computational resources supported by the Academic and Research Computing group at Worcester Polytechnic Institute. 
}


\bibliography{references}

\begin{thebibliography}{39}
\providecommand{\natexlab}[1]{#1}
\providecommand{\url}[1]{\texttt{#1}}
\expandafter\ifx\csname urlstyle\endcsname\relax
  \providecommand{\doi}[1]{doi: #1}\else
  \providecommand{\doi}{doi: \begingroup \urlstyle{rm}\Url}\fi

\bibitem[Chang et~al.(2017)Chang, Meng, Haber, Tung, and
  Begert]{chang2017multi}
B.~Chang, L.~Meng, E.~Haber, F.~Tung, and D.~Begert.
\newblock Multi-level residual networks from dynamical systems view.
\newblock \emph{arXiv preprint arXiv:1710.10348}, 2017.

\bibitem[Chen et~al.(2018)Chen, Rubanova, Bettencourt, and
  Duvenaud]{chen2018neural}
R.~T. Chen, Y.~Rubanova, J.~Bettencourt, and D.~K. Duvenaud.
\newblock Neural ordinary differential equations.
\newblock \emph{Advances in neural information processing systems}, 31, 2018.

\bibitem[Chollet et~al.(2015)]{chollet2015keras}
F.~Chollet et~al.
\newblock Keras.
\newblock \url{https://keras.io}, 2015.

\bibitem[Croitoru et~al.(2023)Croitoru, Hondru, Ionescu, and
  Shah]{croitoru2023diffusion}
F.-A. Croitoru, V.~Hondru, R.~T. Ionescu, and M.~Shah.
\newblock Diffusion models in vision: A survey.
\newblock \emph{IEEE Transactions on Pattern Analysis and Machine
  Intelligence}, 2023.

\bibitem[Falcon and {The PyTorch Lightning
  team}(2019)]{Falcon_PyTorch_Lightning_2019}
W.~Falcon and {The PyTorch Lightning team}.
\newblock {PyTorch Lightning}, 3 2019.
\newblock URL \url{https://github.com/Lightning-AI/lightning}.

\bibitem[Goodfellow et~al.(2016)Goodfellow, Bengio, and
  Courville]{Goodfellow-et-al-2016}
I.~Goodfellow, Y.~Bengio, and A.~Courville.
\newblock \emph{Deep Learning}.
\newblock MIT Press, 2016.
\newblock \url{http://www.deeplearningbook.org}.

\bibitem[Gu and Dao(2023)]{gu2023mamba}
A.~Gu and T.~Dao.
\newblock Mamba: Linear-time sequence modeling with selective state spaces.
\newblock \emph{arXiv preprint arXiv:2312.00752}, 2023.

\bibitem[Gu et~al.(2021)Gu, Goel, and R{\'e}]{gu2021efficiently}
A.~Gu, K.~Goel, and C.~R{\'e}.
\newblock Efficiently modeling long sequences with structured state spaces.
\newblock \emph{arXiv preprint arXiv:2111.00396}, 2021.

\bibitem[Gunther et~al.(2020)Gunther, Ruthotto, Schroder, Cyr, and
  Gauger]{gunther2020layer}
S.~Gunther, L.~Ruthotto, J.~B. Schroder, E.~C. Cyr, and N.~R. Gauger.
\newblock Layer-parallel training of deep residual neural networks.
\newblock \emph{SIAM Journal on Mathematics of Data Science}, 2\penalty0
  (1):\penalty0 1--23, 2020.

\bibitem[Han et~al.(2019)Han, Li, et~al.]{han2019mean}
J.~Han, Q.~Li, et~al.
\newblock A mean-field optimal control formulation of deep learning.
\newblock \emph{Research in the Mathematical Sciences}, 6\penalty0
  (1):\penalty0 1--41, 2019.

\bibitem[He et~al.(2016)He, Zhang, Ren, and Sun]{he2016deep}
K.~He, X.~Zhang, S.~Ren, and J.~Sun.
\newblock Deep residual learning for image recognition.
\newblock In \emph{Proceedings of the IEEE conference on computer vision and
  pattern recognition}, pages 770--778, 2016.

\bibitem[Hershey(2023)]{hershey2023exploring}
Q.~Hershey.
\newblock \emph{Exploring Neural Network Structure through Iterative Neural
  Networks: Connections to Dynamical Systems}.
\newblock PhD thesis, Worcester Polytechnic Institute, 2023.

\bibitem[Hershey et~al.(2023)Hershey, Paffenroth, and Pathak]{hershey2023deep}
Q.~Hershey, R.~C. Paffenroth, and H.~Pathak.
\newblock Exploring neural network structure through sparse recurrent neural
  networks: A recasting and distillation of neural network hyperparameters.
\newblock In \emph{2023 22nd IEEE International Conference On Machine Learning
  And Applications (ICMLA)}. IEEE, 2023.

\bibitem[Huang et~al.(2017)Huang, Liu, Van Der~Maaten, and
  Weinberger]{huang2017densely}
G.~Huang, Z.~Liu, L.~Van Der~Maaten, and K.~Q. Weinberger.
\newblock Densely connected convolutional networks.
\newblock In \emph{Proceedings of the IEEE conference on computer vision and
  pattern recognition}, pages 4700--4708, 2017.

\bibitem[K{\"a}stner and Crook(2023)]{kastner2023explaining}
L.~K{\"a}stner and B.~Crook.
\newblock Explaining ai through mechanistic interpretability.
\newblock 2023.

\bibitem[LeCun et~al.(2010)LeCun, Cortes, Burges, et~al.]{lecun2010mnist}
Y.~LeCun, C.~Cortes, C.~Burges, et~al.
\newblock Mnist handwritten digit database, 2010.

\bibitem[LeCun et~al.(2015)LeCun, Bengio, and Hinton]{lecun2015deep}
Y.~LeCun, Y.~Bengio, and G.~Hinton.
\newblock Deep learning.
\newblock \emph{nature}, 521\penalty0 (7553):\penalty0 436--444, 2015.

\bibitem[Lipton et~al.(2015)Lipton, Berkowitz, and Elkan]{lipton2015critical}
Z.~C. Lipton, J.~Berkowitz, and C.~Elkan.
\newblock A critical review of recurrent neural networks for sequence learning.
\newblock \emph{arXiv preprint arXiv:1506.00019}, 2015.

\bibitem[Liu and Theodorou(2019)]{liu2019deep}
G.-H. Liu and E.~A. Theodorou.
\newblock Deep learning theory review: An optimal control and dynamical systems
  perspective.
\newblock \emph{arXiv preprint arXiv:1908.10920}, 2019.

\bibitem[maintainers and contributors(2016)]{torchvision}
T.~maintainers and contributors.
\newblock {TorchVision: PyTorch's Computer Vision library}, 11 2016.
\newblock URL \url{https://github.com/pytorch/vision}.

\bibitem[Michaud et~al.(2024)Michaud, Liao, Lad, Liu, Mudide, Loughridge, Guo,
  Kheirkhah, Vukeli{\'c}, and Tegmark]{michaud2024opening}
E.~J. Michaud, I.~Liao, V.~Lad, Z.~Liu, A.~Mudide, C.~Loughridge, Z.~C. Guo,
  T.~R. Kheirkhah, M.~Vukeli{\'c}, and M.~Tegmark.
\newblock Opening the ai black box: program synthesis via mechanistic
  interpretability.
\newblock \emph{arXiv preprint arXiv:2402.05110}, 2024.

\bibitem[Miljanovic(2012)]{miljanovic2012comparative}
M.~Miljanovic.
\newblock Comparative analysis of recurrent and finite impulse response neural
  networks in time series prediction.
\newblock \emph{Indian Journal of Computer Science and Engineering}, 3\penalty0
  (1):\penalty0 180--191, 2012.

\bibitem[Murtagh(1991)]{MURTAGH1991183}
F.~Murtagh.
\newblock Multilayer perceptrons for classification and regression.
\newblock \emph{Neurocomputing}, 2\penalty0 (5):\penalty0 183--197, 1991.
\newblock ISSN 0925-2312.
\newblock \doi{https://doi.org/10.1016/0925-2312(91)90023-5}.
\newblock URL
  \url{https://www.sciencedirect.com/science/article/pii/0925231291900235}.

\bibitem[Narang et~al.(2017)Narang, Undersander, and
  Diamos]{narang2017blocksparse}
S.~Narang, E.~Undersander, and G.~Diamos.
\newblock Block-sparse recurrent neural networks, 2017.

\bibitem[Pascanu et~al.(2013)Pascanu, Mikolov, and Bengio]{pmlr-v28-pascanu13}
R.~Pascanu, T.~Mikolov, and Y.~Bengio.
\newblock On the difficulty of training recurrent neural networks.
\newblock In S.~Dasgupta and D.~McAllester, editors, \emph{Proceedings of the
  30th International Conference on Machine Learning}, volume~28 of
  \emph{Proceedings of Machine Learning Research}, pages 1310--1318, Atlanta,
  Georgia, USA, 6 2013. PMLR.
\newblock URL \url{https://proceedings.mlr.press/v28/pascanu13.html}.

\bibitem[Paszke et~al.(2017)Paszke, Gross, Chintala, Chanan, Yang, DeVito, Lin,
  Desmaison, Antiga, and Lerer]{pytorch2017automatic}
A.~Paszke, S.~Gross, S.~Chintala, G.~Chanan, E.~Yang, Z.~DeVito, Z.~Lin,
  A.~Desmaison, L.~Antiga, and A.~Lerer.
\newblock Automatic differentiation in pytorch.
\newblock In \emph{NIPS-W}, 2017.

\bibitem[Pathak et~al.(2023)Pathak, Paffenroth, and Hershey]{pathak2023deep}
H.~Pathak, R.~C. Paffenroth, and Q.~Hershey.
\newblock Sequential2d: Organizing center of skip connections for transformers.
\newblock In \emph{2023 22nd IEEE International Conference On Machine Learning
  And Applications (ICMLA)}. IEEE, 2023.

\bibitem[Radhakrishnan et~al.(2020)Radhakrishnan, Belkin, and
  Uhler]{radhakrishnan2020overparameterized}
A.~Radhakrishnan, M.~Belkin, and C.~Uhler.
\newblock Overparameterized neural networks implement associative memory.
\newblock \emph{Proceedings of the National Academy of Sciences}, 117\penalty0
  (44):\penalty0 27162--27170, 2020.

\bibitem[Rasamoelina et~al.(2020)Rasamoelina, Adjailia, and Sinčák]{9108717}
A.~D. Rasamoelina, F.~Adjailia, and P.~Sinčák.
\newblock A review of activation function for artificial neural network.
\newblock In \emph{2020 IEEE 18th World Symposium on Applied Machine
  Intelligence and Informatics (SAMI)}, pages 281--286, 2020.
\newblock \doi{10.1109/SAMI48414.2020.9108717}.

\bibitem[Rumelhart and McClelland(1987)]{rnn}
D.~E. Rumelhart and J.~L. McClelland.
\newblock \emph{Learning Internal Representations by Error Propagation}, pages
  318--362.
\newblock MIT Press, 1987.

\bibitem[Salehinejad et~al.(2017)Salehinejad, Sankar, Barfett, Colak, and
  Valaee]{salehinejad2017recent}
H.~Salehinejad, S.~Sankar, J.~Barfett, E.~Colak, and S.~Valaee.
\newblock Recent advances in recurrent neural networks.
\newblock \emph{arXiv preprint arXiv:1801.01078}, 2017.

\bibitem[Sch{\"a}fer and Zimmermann(2006)]{schafer2006recurrent}
A.~M. Sch{\"a}fer and H.~G. Zimmermann.
\newblock Recurrent neural networks are universal approximators.
\newblock In \emph{Artificial Neural Networks--ICANN 2006: 16th International
  Conference, Athens, Greece, September 10-14, 2006. Proceedings, Part I 16},
  pages 632--640. Springer, 2006.

\bibitem[Shen et~al.(2023)Shen, Jin, Huang, Liu, Dong, Guo, Wu, Liu, and
  Xiong]{shen2023large}
T.~Shen, R.~Jin, Y.~Huang, C.~Liu, W.~Dong, Z.~Guo, X.~Wu, Y.~Liu, and
  D.~Xiong.
\newblock Large language model alignment: A survey.
\newblock \emph{arXiv preprint arXiv:2309.15025}, 2023.

\bibitem[Sherstinsky(2020)]{Sherstinsky_2020_rnn}
A.~Sherstinsky.
\newblock Fundamentals of recurrent neural network ({RNN}) and long short-term
  memory ({LSTM}) network.
\newblock \emph{Physica D: Nonlinear Phenomena}, 404:\penalty0 132306, mar
  2020.
\newblock \doi{10.1016/j.physd.2019.132306}.

\bibitem[Siegelmann and Sontag(1991)]{siegelmann1991turing}
H.~T. Siegelmann and E.~D. Sontag.
\newblock Turing computability with neural nets.
\newblock \emph{Applied Mathematics Letters}, 4\penalty0 (6):\penalty0 77--80,
  1991.

\bibitem[Weinan(2017)]{weinan2017proposal}
E.~Weinan.
\newblock A proposal on machine learning via dynamical systems.
\newblock \emph{Communications in Mathematics and Statistics}, 1\penalty0
  (5):\penalty0 1--11, 2017.

\bibitem[Yang et~al.(2023)Yang, Zhang, Song, Hong, Xu, Zhao, Zhang, Cui, and
  Yang]{yang2023diffusion}
L.~Yang, Z.~Zhang, Y.~Song, S.~Hong, R.~Xu, Y.~Zhao, W.~Zhang, B.~Cui, and
  M.-H. Yang.
\newblock Diffusion models: A comprehensive survey of methods and applications.
\newblock \emph{ACM Computing Surveys}, 56\penalty0 (4):\penalty0 1--39, 2023.

\bibitem[Yeung et~al.(2019)Yeung, Kundu, and Hodas]{yeung2019learning}
E.~Yeung, S.~Kundu, and N.~Hodas.
\newblock Learning deep neural network representations for koopman operators of
  nonlinear dynamical systems.
\newblock In \emph{2019 American Control Conference (ACC)}, pages 4832--4839.
  IEEE, 2019.

\bibitem[Zhang et~al.(2021)Zhang, Ti{\v{n}}o, Leonardis, and
  Tang]{zhang2021survey}
Y.~Zhang, P.~Ti{\v{n}}o, A.~Leonardis, and K.~Tang.
\newblock A survey on neural network interpretability.
\newblock \emph{IEEE Transactions on Emerging Topics in Computational
  Intelligence}, 5\penalty0 (5):\penalty0 726--742, 2021.

\end{thebibliography}


\end{document}